\documentclass{article}

\usepackage{ijcai21}

\usepackage{natbib}
\usepackage{times}
\usepackage{latexsym}
\usepackage{graphicx} % For including images
\usepackage{subcaption} % For subfigures
\usepackage{setspace}
\usepackage{microtype}

\usepackage{inconsolata}

\usepackage{float}
\usepackage[utf8]{inputenc} % allow utf-8 input
\usepackage[T1]{fontenc}    % use 8-bit T1 fonts
\usepackage{hyperref}       % hyperlinks
\usepackage{url}            % simple URL typesetting
\usepackage{booktabs}       % professional-quality tables
\usepackage{amsfonts}       % blackboard math symbols
\usepackage{nicefrac}       % compact symbols for 1/2, etc.
\usepackage{microtype}      % microtypography
\usepackage{lipsum}
\usepackage{fancyhdr}       % header
\usepackage{graphicx}       % graphics

\usepackage{multirow}       % needed for table
\usepackage{makecell}       % needed for table

\usepackage{amsfonts}
\usepackage{amsmath,amssymb}
\usepackage{amsthm}
\usepackage{graphicx}
\usepackage{newfloat}
\usepackage{listings}
\DeclareCaptionStyle{ruled}{labelfont=normalfont,labelsep=colon,strut=off} % DO NOT CHANGE THIS
\floatstyle{ruled}
\newfloat{listing}{tb}{lst}{}
\floatname{listing}{Listing}

\newtheorem{example}{Example}
\newtheorem{property}{Property}
\newtheorem{definition}{Definition}
\newtheorem{proposition}{Proposition}
\newtheorem{notation}{Notation}

\newcommand{\Args}{\mathcal{A}}
\newcommand{\Atts}{\mathcal{R}^-}

\newcommand{\Supps}{\mathcal{R}^+}
\newcommand{\BS}{\ensuremath{\tau}}
\newcommand{\SF}{\ensuremath{\sigma}}

\newcommand{\BAF}{\mathcal{B}}
\newcommand{\QBAF}{\mathcal{Q}}
\newcommand{\Pros}{\ensuremath{\mathsf{pro}}}
\newcommand{\Cons}{\ensuremath{\mathsf{con}}}
\newcommand{\argpaths}{\mathsf{paths}}
\usepackage{todonotes}
\usepackage{xargs}
\usepackage{xcolor}

\graphicspath{{images}}

\long\def\symbolfootnote[#1]#2{\begingroup%
\def\thefootnote{\fnsymbol{footnote}}\footnote[#1]{#2}\endgroup}

\makeatother

\newenvironment{links}{%
  \newcommand{\link}[3]{\par\textbf{##1} --- \href{##3}{##2}}%
  \setlength{\hangindent}{10pt}%
  \setlength{\parskip}{2pt}%
  \begin{flushleft}%
}{%
  \end{flushleft}%
  \vskip 1ex%
}%
\title{Argumentative Large Language Models for \\ Explainable and Contestable Claim Verification}

\author{Gabriel Freedman* \and Adam Dejl*  \and Deniz Gorur* \and Xiang Yin* \and Antonio Rago \and Francesca Toni  \\
  Department of Computing, Imperial College London, UK \\
  \texttt{\{gif22, adam.dejl18, d.gorur22, xy620, a.rago, ft\}@imperial.ac.uk}
  }

\begin{document}

\maketitle

\let\thefootnote\relax\footnotetext{Accepted as an oral presentation at AAAI 2025.}

\begin{abstract}
The %diversity of 
profusion of knowledge encoded in large language models (LLMs) and their ability to apply this knowledge zero-shot in a range of settings makes them promising candidates for use in decision-making. However, they are currently limited by their inability to provide outputs which 
%are explainable and %are not
%contestable 
can be faithfully explained and effectively contested to correct  mistakes%(i.e. their mistakes cannot be reliably corrected)
.
% and can be reliably contested (and corrected)%(i.e. contested)  by human usersLET US NOT MENTION HUMANS TO MANAGE EXPECTATIONS: NO USER STUDY. also, why not contesdtability by AI? why exclude it?
In this paper, we attempt to reconcile these strengths and weaknesses by introducing \emph{argumentative LLMs (ArgLLMs)},
a method for augmenting %\todo{augmenting? extending? enriching?}
LLMs with argumentative reasoning. Concretely, 
%we introduce \emph{argumentative LLMs}, a method utilising LLMs to
ArgLLMs construct argumentation frameworks, which then serve as the basis for formal reasoning in support of decision-making. The interpretable nature of these argumentation frameworks and formal reasoning  means that any decision made by 
%the supplemented LLM 
ArgLLMs may be %naturally 
explained %to, 
and contested% by, humans LET US NOT MENTION HUMANS TO MANAGE EXPECTATIONS: NO USER STUDY
. We %demonstrate the effectiveness of 
evaluate ArgLLMs' performance experimentally in comparison with state-of-the-art techniques, %using
in the context of the decision-making task of claim verification. %We obtain accuracy scores that are competitive with, and in some cases surpass, comparable state-of-the-art techniques, while also delivering explainable and contestable outputs.
We also define novel properties to characterise contestability and assess %the contestable nature of 
ArgLLMs formally in terms of these properties.

% \note{might need to add a bit more on the experimental details but maybe not since it's already long}
\end{abstract}

 \begin{links}
     \link{Code and Datasets}{github.com/CLArg-group/argumentative-llms}{https://github.com/CLArg-group/argumentative-llms}
     % \link{Datasets}{https://aaai.org/example/datasets}
     % \link{Extended version}{https://arxiv.org/abs/2405.02079}
\end{links}
% keywords can be removed

\section{Introduction}
\label{sec:introduction}

\begin{figure}[tb!]
    \centering
    % \hspace{0.1cm}
    \begin{minipage}{0.4\textwidth}
        \centering
        % \hspace{-4cm} \includegraphics[width=\textwidth]{AAAI/baselines.pdf}
         \includegraphics[width=\textwidth]{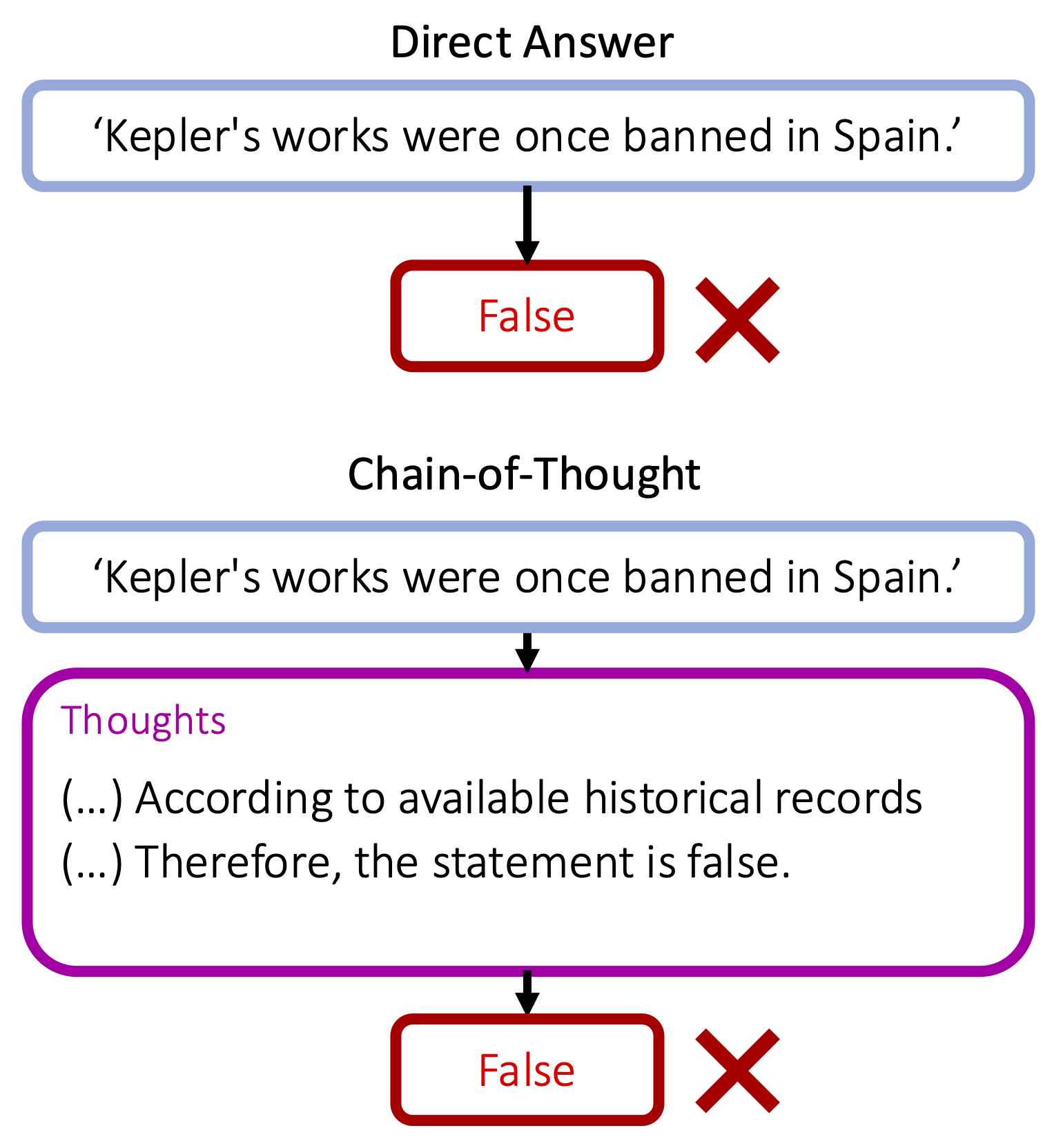}
    \end{minipage}

   \vspace{0.1cm}
    
    \begin{minipage}{0.45\textwidth}
        \centering
        % % \hspace{-5.25cm} \includegraphics[width=\textwidth]{AAAI/argu_llms.pdf}
        \includegraphics[width=\textwidth]{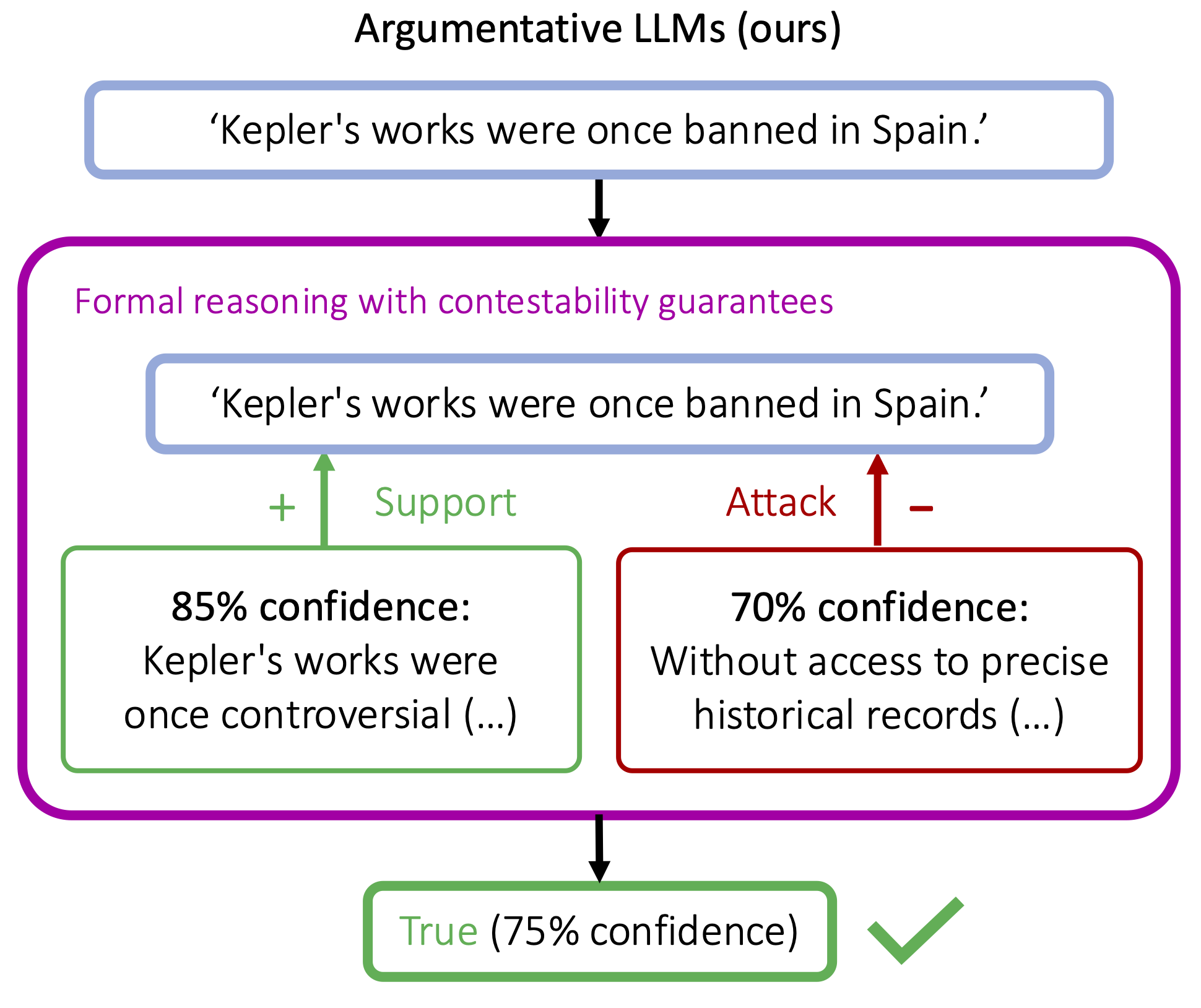}
    \end{minipage}

    \caption{Comparison of our approach (\emph{ArgLLM}, here in combination with Mixtral) with existing alternatives. The example claim is adapted from TruthfulQA%(TruthfulClaim) 
    %and the %abridged 
    %outputs are generated by Mixtral 
    .}
    \label{fig:example}
\end{figure}

The profusion of knowledge encoded in large language models (LLMs) and their ability to apply this knowledge zero-shot in a range of settings (e.g. as in \citet{Brown2020LanguageMA, Bubeck2023SparksOA,gpt-4-tech-report}) makes them 
%Large language models (LLMs) have produced excellent results on a diverse range of reasoning tasks \cite{Brown2020LanguageMA, Bubeck2023SparksOA}. This capacity has made them compelling 
promising candidates for supporting automated decision making \cite{Zhang2023IntegratingAK,Ouyang2023AutoPlanAP,Wang2023RCAgentCR}.
However, they are currently limited by their inability to explain their outputs faithfully, i.e. in a way that reflects their ``reasoning'' and %the knowledge used
knowledge. LLMs also lack contestability, meaning that there is no mechanism for external agents to reliably dispute and correct the reasoning steps taken by the model: although users may attempt contestation through prompting, the highly stochastic nature of LLMs provides no guarantee of this achieving the intended outcome.
These abilities (to be explainable and contestable) are important for AI systems in general~\cite{Henin2021BeyondEJ,Lyons2021ConceptualisingC,ourKRcontestabilitypaperARXIV} and for LLMs in particular, given 
their %reasoning abilities currently suffer from various 
limitations, e.g. hallucinations and logical inconsistencies \cite{Shanahan24, berglund2023reversal, fluri2023evaluating}% Deficiencies which are particularly worrying are a lack of explainability and inability to provide faithful representations of their reasoning, 
%which raise questions regarding their trustworthiness% and ability to be contested by humans regarding their decisions~\cite{Henin2021BeyondEJ,Lyons2021ConceptualisingC}
.
In this paper 
we explore the following question, grounded in the context of claim verification as a form of decision-making:

\begin{quote}
Can LLMs be augmented with the ability to construct formal arguments in order to give explainable and contestable outputs?
\end{quote}
% Can the reasoning abilities of LLMs improve if they are %prompted 
% made to argue with themselves? 
%Can the reasoning process of LLMs be improved upon by using their outputs to construct a formal argument?

 \begin{figure*}[!tb]
\centering
\includegraphics[scale=0.085]{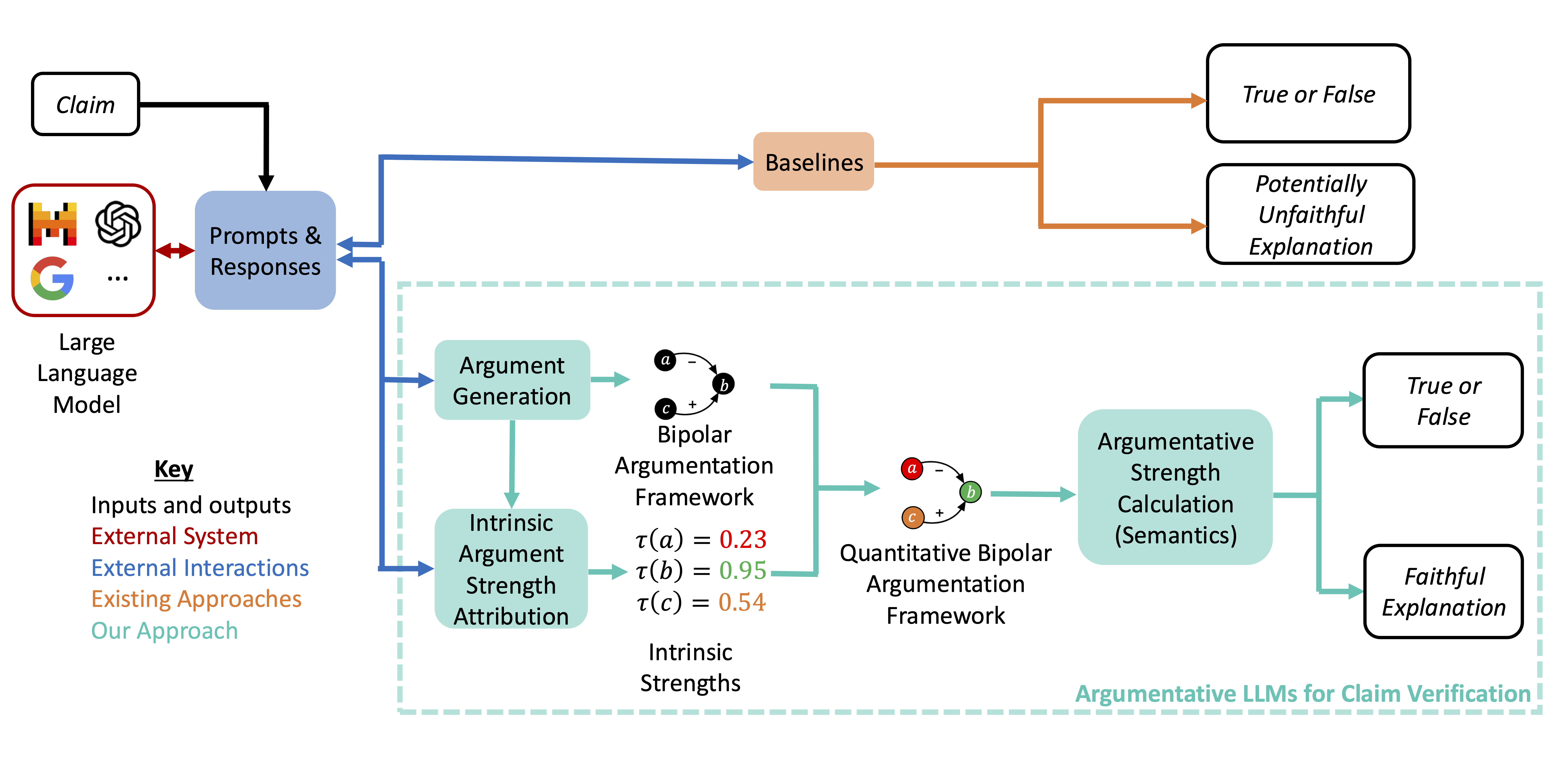}
\caption{\label{fig:pipeline} Pipeline for ArgLLMs %for claim verification 
(%alongside 
in comparison with %the spirit of
baselines, %focusing on the claim verification task
see \S\ref{sec:main} and \S\ref{sec:experiments} for the details).
}
\end{figure*}

\noindent The question is inspired by argumentative interpretations of human reasoning~\cite{argtheory,mercier2018enigma} and by 
the fact that formal argumentation (as understood in~\citet{AImagazine17}) has been shown to excel in supporting decision making~\cite{arg-dec-2009} and explainability~\cite{ArgXAIsurvey}. It is also advocated as a suitable approach towards contestability~\cite{ourKRcontestabilitypaperARXIV}.
% A positive answer to this question would have implications as regards both explainability~\cite{} and contestability~\cite{disputableXAI}.
Further, \citet{hammond2023large} argue that the inherent limitations of LLMs can be remedied by the use of symbolic AI techniques, which include formal argumentation.
In a nutshell,  this (also referred to as `computational argumentation'), 
represents information
in terms of `arguments' and dialectical relations (of `attack'
and, possibly, `support') between them; it is equipped with
`semantics' (and algorithms) to reach %some form of 
consensus %regarding 
on conclusions to be formally drawn.

% \begin{figure*}[htp!]
%     \centering
%     \includegraphics[width=0.7\textwidth]{AAAI/arg_example.pdf}
%     \caption{Comparison of our approach (\emph{ArgLLM}) with existing alternatives. The example claim is adapted from TruthfulQA (TruthfulClaim) and the abridged outputs are generated by Mixtral. TODO: MAKE THIS VERTICAL AND IN THE FIRST PAGE}
%     \label{fig:example}
% \end{figure*}

Concretely, we propose \emph{argumentative LLMs (ArgLLMs)}. %alongside baselines, focusing on the claim verification task.
While existing approaches \cite{chain-of-thought2022,tree-of-thought23} prompt LLMs to produce `thoughts' that either enrich the context of the LLMs or provide disparate reasoning steps% to compare
,
% Rather than prompting an LLM to produce `thoughts', as in \cite{chain-of-thought2022} or \cite{tree-of-thought23}, that either enrich the context of the LLM or provide disparate reasoning steps% to compare
ArgLLMs %can be seen as providing
provide %`thoughts' 
`arguments' for and against particular outputs, in the spirit of \citet{evaluativeAI-Miller23}, as illustrated for claim verification in Figure~\ref{fig:example}. This feature of ArgLLMs makes them a natural fit for %highly
complex decision-making tasks (such as claim verification), wherein an option %, or set of options, 
must be chosen from a number of possible alternatives. Indeed, %almost all 
in many real-world settings, a particular decision will have both pros and cons, which %is a feature that 
ArgLLMs %both 
formalise and leverage.  

Overall, we aim to achieve a broader `improvement' in the LLMs' reasoning%
%we take a broader than usual view of what counts as an %`improvement' for %reasoning process
%LLMs
, in the spirit of~\citet{liao2024ai}:
%. 
in addition to 
%demonstrating that our \emph{ArgLLMs} 
achieving %competitive scores
reasonable performance%  in claim verification
, %on various reasoning benchmarks, we show how an improved reasoning process necessarily leads to more explainable and contestable decision-making~\cite{liao2024ai}.
we target explainability and 
contestability.
While existing methods for improving the reasoning of LLMs (such as \citet{chain-of-thought2022,lewis2020retrieval}) do not necessitate a direct relationship between the reasoning steps and the final decision \cite{turpin2024language}, our argumentative approach %, on the other hand, 
provides this as a feature of the system. 
%CHECK: This is because our approach can be seen as introducing a kind of local, explainable `surrogate' models \cite{BlancoJusticia2019 MachineLE, Nbrega2019TowardsER}, in the form of argumentation frameworks \cite{Bentahar2010ATO}. 
% Instead of training an interpretable model on the input-output pairs of a black-box model, 
Indeed, we use the outputs of LLMs to perform deterministic inference under \emph{gradual semantics} with \emph{quantitative bipolar argumentation frameworks} (QBAFs)~\cite{Baroni_19}. %These frameworks 
QBAFs comprise of arguments, each equipped with an \emph{intrinsic strength}, and attack and support relations. They are inherently interpretable and thus, as stated in \citet{rudin2019stop}, ``%models that are inherently interpretable... 
provide their own explanations, which are faithful to what the model actually computes''. %\note{This is analogous to how the rules forming a decision tree necessarily constitute a faithful explanation of a decision-tree model.}
This is analogous to how the rules forming a decision-tree model necessarily constitute a faithful explanation of the model.

%This means that, because the surrogate model (argumentation framework) we use to determine the final output is interpretable, the model itself can be used as a faithful explanation of the output.
%

Further, ArgLLMs also provide a guarantee of contestability, in that %if a human intervenes 
an intervention in the reasoning process by modifying the QBAF
(such as by adding %or removing an argument
support for an argument 
% or removing an existing one, 
or by changing its intrinsic strength) %this 
will have a measurable effect on the output% of the 
% decision-making system.
%system
.
%Comparable techniques lack the necessary processing stage between the LLM's output and the final decision to accommodate this flexibility.

Throughout the paper, we focus on %the kind of reasoning underpinning  
claim verification as the decision-making task of interest. This setting lends itself well to %our framework 
ArgLLMs as claims are often under-determined, so they %do not necessarily 
may not 
have straightforward truth values. By intrinsically considering both arguments in favour of and in conflict with the truthfulness of claims, ArgLLMs are able to ascertain the best answer given the available evidence.
For simplicity, and without loss of generality,\footnote{ArgLLMs can be easily extended to the case where there are multiple possible options by generating a set of candidate answers and applying %the 
our method to each of these.} we focus on the binary setting, rather than general question-answer% problems 
ing as in \citet{chain-of-thought2022} and \citet{tree-of-thought23}. %In order to handle open-ended settings, one can first generate candidate answers — where determining the optimal number of answers can be thought of as a hyperparameter. A DETAIL TO BE ADDED WHEN WE TALK OF DATA GENERATION....

In summary, we make the following contributions:
\begin{itemize}
    \item We define ArgLLMs, a novel interpretable method %for
    %supplementing 
    augmenting
    %\todo{use same term as in the absrtact}
    LLMs with formal, argumentative reasoning.% for decision-making
    \item We perform an 
    % extensive 
    evaluation of ArgLLMs' %decision-making abilities for claim verification 
    claim verification abilities by comparing four variants thereof with three baselines (two based on direct prompting, plus the chain-of-thought approach~\cite{chain-of-thought2022}), on three novel claim verification datasets, adapted from existing datasets (TruthfulQA~\cite{truthfulqa21}, StrategyQA~\cite{strategyqa21} and MedQA~\cite{medqa20}). %
    The evaluation shows that ArgLLMs %perform %as well as, and sometimes better than, 
    deliver performance comparable to the baselines with the added benefit of being faithfully explainable. 
    %comparably as the baselines, while being faithfully explainable.
    \item We formally characterise contestability properties for %decision-making systems and nstantiate them to the setting of 
    ArgLLMs, %prove formally demonstrate the explainability and contestability benefits of ArgLLMs.
    showing formally in which sense ArgLLMs may be deemed contestable. 
\end{itemize}
Figure~\ref{fig:pipeline} gives an overview of our approach, which can be deployed in combination with any LLM.
%
% \todo{\cite{Cyras_21,Vassiliades_21,Guo_23}}

% \begin{figure*}
% % \includegraphics[width=\linewidth]{../images/pipeline.png}
% \includegraphics[scale=0.2]{../images/pipeline.png}
% \caption{\label{fig:pipeline} Pipeline for Argumentative LLMs (%alongside 
% in comparison with %the spirit of
% baselines).  %\FT{TO DO: HAVE A SINGLE GENERIC "BASELINE" AT THE TOP} \FT{ALSO: put ...alongside the LLMs on the left? \note{Change Argument Mining -> Argument Generation, Uncertainty Estimation -> Argument Strength Attribtution. Add an arrow from Argument Generation to Arg Strength Attribution. change meta to ...}}
% }
% \end{figure*}

% Background Section?

\section{Related Work}
\label{sec:related}

Existing research into LLM optimisation has primarily focused on improving performance on reasoning benchmarks, with improvements in explainability treated as a desirable byproduct. This existing research can be coarsely divided into approaches which exclusively focus on prompt-engineering \cite{chain-of-thought2022,Yang2023LargeLM}, and those which endow LLMs with the ability to utilise external tools, or information, or extra structural constraints \cite{schick2024toolformer, tree-of-thought23, lewis2020retrieval}. Our method is more closely aligned to the latter, %as DOES IT FOLLOW? it results 
resulting in symbolic, deterministically evaluable %graphs 
%argumentation frameworks 
arguments as its output.

% \citet{du2023improving} also use arguments to improve the reasoning ability of LLMs. However, they focus on a multi-agent setting and do not formalise the arguments produced by LLMs, or their corresponding strengths. Also at the intersection of LLMs and argumentation is work looking at the efficacy of LLMs at completing arguments \cite{thorburn2022optimizing}, and the persuasiveness of LLM generated arguments \cite{hinton2023persuasive,durmus2024persuasion}.

\emph{Chain-of-thought} approaches \cite{chain-of-thought2022, zhang2023chain} attempt to induce enhanced reasoning through a specific form of prompting. The prompt specifies (using either few-shot examples or a verbal description) that the problem should be broken down into discrete steps, before the final decision is outputted. However, all the reasoning takes place within the autoregressively generated output of the model. Due to the nature of the next token prediction mechanism underlying these models, this does not guarantee that the steps in the reasoning, or the final output, actually follow from each other \cite{turpin2024language,xia2024evaluating}. This undermines the premise that the reasoning is faithful to the process taking place in the model or that it is directly related to the final output.

\citet{tafjord2022entailer} train an extra model to measure entailment between reasoning steps and the final output. While this provides some level of assurance that the output will be causally related to the explanation, it relies on an external opaque model, which cannot provide any guarantees. %In comparison to this, 
Instead,
our method  builds an argumentation framework which guarantees a resolution solely based on the constituent entities, offering the added advantages of innately dealing with conflict and being amenable to contestation.

Other related approaches are \emph{tree-of-thought} \cite{tree-of-thought23} and \emph{graph-of-thoughts} \cite{besta2024graph}. Similarly to our method, %these approaches 
they result in graph-like structures, composed of the LLM's output, which can then be reasoned over post-hoc. In contrast to our method, the nodes of these graphs consist of decomposed components of the overall problem. Instead, %our method permits 
ArgLLMs permit a comprehensive and %fully 
explainable reasoning process to take place concerning a single claim, which may be controversial or highly complex%, in addition to composite problems like the %alternate 
%existing methods
. For example, a claim such as `it is a good idea to drink milk when you have a cough', does not naturally lend itself to decomposition, but would benefit from argumentative reasoning, i.e. evaluating arguments for (e.g. `it is a traditional remedy') and against (e.g. `there have been no scientific studies confirming this'). In addition, to the best of our knowledge, no %above methodologies, and indeed any others that are not present, \delete{do no} 
existing approach attempts to integrate any formal guarantees of contestability% in contrast to our methodology
, which ArgLLMs naturally provide.

% nice

% Our approach can be seen as prompting LLMs to explain their answers argumentatively. \FT{THE BELOW COULD GO, AS THE ARTICLE NOT PUBLISHED YET...MAY BE CARTESNS TONI 2017 INSTEAD?} \note{What paper is this referring to exactly?}
% The usefulness of argumentative explanations is advocated in \citet{neema-argX24}, alongside several properties that they can be asked to satisfy, but for explaining fact-checking NLP models. In the context of the general proposal of \citet{neema-argX24}, 
% the argumentative explanations we focus on are of a restricted form - at a maximum, with one base attacker and one base supporter with an additional argument attacking and supporting the base arguments respectively.
% Explainable fact-checking in NLP has been the focus of considerable attention (e.g. see overview in \cite{neema-survey20}). Amongst several approaches, \cite{explainFactChecking-QA22} advocates using question answering for fact-checking, but not in the context of LLMs, and not with the kinds of argumentative explanations we have in mind.

%\todo[inline]{add works extracting logical formulas from text using LLMs??? }

LLMs have also been used to extract logical formulas from text \cite{Ishay_23}, by combining LLMs' prowess in dealing with natural language with the complex reasoning capabilities of reasoning by answer set programming. %They find that complex answer set programs can be generated for logic puzzles, where the errors involved can be easily corrected by humans. This approach differs from ours 
ArgLLMs differ from this approach in that we 
extract symbolic representations directly from LLMs (rather than from externally provided text) and deploy a %quantitative %, dialectical, rather than strictly logical, 
gradual approach to reasoning (rather than a model-theoretic one), with the potential to handle uncertainty.

Also relevant to our work are approaches focused specifically on automated claim verification using LLMs, such as FOLK \cite{wang-claim-verification} or HiSS \cite{zhang-fact-verification}. However, in contrast with ArgLLMs, these works rely on access to external sources or knowledge bases.

\section{Preliminaries}
\label{sec:preliminaries}

%{\bf Claim Verification}
\paragraph{Claim Verification}
%The considered task of claim verification can be divided into two primary types: \todo{can we add a reference here?} 
We consider two types of claim verification: \textit{unconditioned} and \textit{conditioned}. For \textit{unconditioned} verification, a claim $c$ is evaluated independently, without any contextual information. The outcome of this evaluation is binary, represented as $v(c) \in \{0, 1\}$, where $v(c) = 1$ denotes the claim is true, and $v(c) = 0$ %indicates 
denotes the claim is false.
Meanwhile, \textit{conditioned} verification considers a claim $c$ given additional information or context $i$ (with the context assumed to be truthful). The veracity of %this tuple 
$c\mid i$ is also assessed in a binary manner, expressed as $v(c \mid i) \in \{0, 1\}$. Here, similarly, $v(c \mid i) = 1$ %signifies that 
denotes the claim $c$, given the context $i$, is true, while $v(c \mid i) = 0$ %means 
denotes it is false. For ease of reference, we use $x$ to denote either $c$ or $c \mid i$.

\paragraph{Computational Argumentation}
%{\bf Computational Argumentation}
A QBAF \cite{Baroni_19} is a quadruple $\langle \Args, \Atts, \Supps, \BS \rangle$ comprising: a set of \emph{arguments} $\Args$; binary, directed relations of \emph{attack} $\Atts\!\subseteq\! \Args \!\times \!\Args$ and \emph{support} $\Supps\!\subseteq\! \Args \!\times \!\Args$, where $\Atts \!\cap \!\Supps = \emptyset$; and a total function $\BS \!: \!\Args \!\rightarrow\! [0,1]$, where for any $\alpha \in \Args$, 
$\BS(\alpha)$ is the \emph{base score} of $\alpha$.\footnote{%Note that the 
The codomain of %the base score 
$\BS$ is defined more generally by \citet{Baroni_19}, but we %restrict to 
use here its most common form.} 
For any argument $\alpha \in \Args$, we use $\Atts(\alpha) \!= \!\{ \beta \!\in \! \Args | (\beta, \alpha) \!\in \! \Atts \}$ to refer to the \emph{attackers} of $\alpha$ and 
$\Supps(\alpha) \!= \!\{ \beta \!\in\! \Args | (\beta, \alpha) \!\in\! \Supps \}$ to refer to the \emph{supporters} of $\alpha$.
%
% For any argument $\alpha \!\in\! \mathcal{A}$, we use $\mathcal{O}(\alpha)\!=\!\{ (\alpha,\beta) \!\in \!\Atts \!\cup\!\Supps \!\mid \! \beta \!\in\! 
% \mathcal{A} \}$ to refer to the set of all %outgoing edges from 
% arguments attacked or supported by $\alpha$.\todo{do we need this?}
Arguments in QBAFs may be evaluated by a \emph{gradual semantics}~\cite{Baroni_19}, i.e. a total function $\SF: \Args \rightarrow [0,1]$ which, 
for any $\alpha \in \Args$, assigns a \emph{strength} $\SF(\alpha)$ to $\alpha$.\footnote{As with %the base score
$\BS$, we use the most %commonly occurring 
common codomain for %gradual semantics
$\SF$.}
While $\BS(\alpha)$ can be seen as the \emph{intrinsic strength} for $\alpha$, the strength $\SF(\alpha)$ can be seen as `dialectical', following the debate captured by $\Atts$ and $\Supps$.
If we ignore $\BS$ in a QBAF, we obtain a \emph{bipolar argumentation framework (BAF)}~\cite{Cayrol_05}.
% \todo{shall we make the arguments here greek like the rest of it?}
%For any set of arguments $S \subseteq \Args$, we denote $\SF(S)$ a sequence (in any order) of all elements of  the set $\{\SF(\argz_i) | \argz_i \in S \}$ (thus $\SF(S)\in \Ione^*$, where $\Ione^*$ is the set of all sequences of elements of $\Ione$).
% When comparing two sequences $\SF(S_1)$ and $\SF(S_2)$, for $S_1,S_2 \subseteq \Args$, we will treat the sequences as multi-sets.

One gradual semantics, the \emph{discontinuity-free quantitative argumentation debate} (DF-QuAD) algorithm \cite{Rago_16}, is such that, for a given QBAF $\langle \Args, \Atts, \Supps, \BS \rangle$, for any $\alpha \in \Args$ with $n\!\geq \! 0$  attackers with strengths $v_1, \ldots, v_n$, $m\geq 0$  supporters with strengths $v_1', \ldots, v_m'$ and $\BS(\alpha)=v_{0}$, $\SF(\alpha)=\mathcal{C}(v_{0}, \mathcal{F}(v_{1}, \ldots, v_{n})$, $\mathcal{F}(v_{1}', \ldots, v_{m}'))$, where
$\mathcal{C}$ and  $\mathcal{F}$ are defined as follows. For $v_{a}=\mathcal{F}(v_{1}, \ldots, v_{n})$ and $v_{s}=\mathcal{F}(v_{1}', \ldots, v_{m}')$:
if $v_a=v_s$ then $\mathcal{C}(v_{0}, v_{a}, v_{s})=v_0$; else if $v_a > v_{s}$ then $\mathcal{C}(v_{0}, v_{a}, v_{s})=v_{0} - (v_{0}\cdot|v_{s}-v_{a}|)$; otherwise $\mathcal{C}(v_{0}, v_{a}, v_{s})=v_{0} + ((1-v_{0})\cdot|v_{s}-v_{a}|)$.
Given $n$ arguments with strengths $v_{1}, \ldots, v_{n}$, if $n=0$ then $\mathcal{F}(v_{1}, \ldots, v_{n})= 0$, otherwise $\mathcal{F}(v_{1}, \ldots, v_{n})= 1 - \prod_{i=1}^n (|1-v_{i}|)$.
Note that DF-QuAD (like all other gradual semantics) is deterministic.

In this paper, we assume all (Q)BAFs are  restricted, %as %follows
%in 
similarly to \cite{Rago_23}, as follows.
%
%\begin{definition}%[(Q)BAFs for arguments]
\label{def:tree}
For $\mathcal{Q}\!=\!\langle \mathcal{A}, \mathcal{R}^{-}, \mathcal{R}^{+}, \tau \rangle$ a QBAF and $\mathcal{B}\!=\!\langle \mathcal{A}, \mathcal{R}^{-}, \mathcal{R}^{+} \rangle$ a BAF, for any $\alpha, \beta \!\in \!\Args$, let a \emph{path} from $\alpha$ to $\beta$ be %defined as
$p\!=\!\langle(\alpha_0,\alpha_1), \ldots, (\alpha_{n-1}, \alpha_{n})\rangle$ for some $n\!>\!0$ (referred to as the \emph{length} of %the path
$p$, denoted $|p|$) where $\alpha_0 = \alpha$, $\alpha_n = \beta$ and, for any $1 \leq i \leq n$, $(\alpha_{i-1}, \alpha_{i}) \in \Atts \cup \Supps$. Let $\argpaths(\alpha,\beta)$ and $|\argpaths(\alpha,\beta)|$ indicate the set of all paths from $\alpha$ to $\beta$,
and the number of paths in $\argpaths(\alpha,\beta)$, respectively.
% Also, we may see paths as sets of pairs.
%
%\todo{Update definition (see first review)}
%\todo{XY:done.}
%
Then, for $\alpha^* \in \Args$,
$\QBAF/\BAF$  is a \emph{QBAF/BAF for $\alpha^*$} iff 

    i) $\forall \alpha \in \mathcal{A}$, $\argpaths(\alpha^*, \alpha)=\emptyset$ and $\argpaths(\alpha, \alpha)=\emptyset$;
    % \nexists (\alpha^*,\alpha) \in \Atts \!\cup \! \Supps$;

    ii) $\forall \alpha \!\in \!\Args \setminus \{\alpha^*\}$,
    $|\argpaths(\alpha,\alpha^*)|=1$.
    % there is %one and only 
    % exactly one path from $\alpha$ to $\alpha^*$; %and 
    % iii) \XY{$\forall \alpha \in \mathcal{A}$, $\argpaths(\alpha, \alpha)=\emptyset$.}
    % $\nexists \alpha \!\in\! \Args$  with a path from $\alpha$ to $\alpha$.
%\end{definition}
\\
Intuitively, (Q)BAFs for $\alpha^*$ can be seen as trees with root $\alpha^*$.
We call \emph{leaves} any unattacked and unsupported arguments in (Q)BAFs  (i.e. $\alpha$ is a leaf if $\Atts(\alpha) \!\cup\! \Supps(\alpha)\!=\!\emptyset)$.

We also use \emph{pro} and \emph{con} arguments in QBAFs 
as in \citet{Rago_23}.
%
%\begin{definition}%[Pro and Con Arguments]
\label{def:procon}
Let $\QBAF$ be a QBAF for $\alpha^*$. Then, the
\emph{pro arguments} and \emph{con arguments} for $\QBAF$ are, respectively:

 $\bullet$ $\Pros(\QBAF) \!\!=\!\! \{ \alpha \!\!\in\!\! \Args | \exists p \!\in\! \argpaths(\alpha,\alpha^*), \text{with} | p \cap \Atts | \text{even} \}$;
 
$\bullet$ $\Cons(\QBAF) \!\!= \!\!\{ \alpha \!\!\in\!\! \Args | \exists p \!\in\! \argpaths(\alpha,\alpha^*), \text{with} | p \cap \Atts | \text{odd} \}$.
In the remainder, unless specified otherwise, we let 
%$\mathcal{B}=\langle \mathcal{A}, \mathcal{R}^{-}, \mathcal{R}^{+} \rangle$, $\mathcal{B}'=\langle \mathcal{A}', \mathcal{R^-}', \mathcal{R^+}' \rangle$,
%\todo{deleted BAFs, where do we need them? }
$\mathcal{Q}$ %and $\mathcal{Q}'=\langle \mathcal{A}', \mathcal{R^-}', \mathcal{R^+}', \tau' \rangle$ denote two (different) QBAFs
denote the QBAF $\langle \mathcal{A}, \mathcal{R}^{-}, \mathcal{R}^{+}, \tau \rangle$. For any gradual semantics $\sigma$, we let 
$\sigma_{\QBAF}(\alpha)$ denote the strength of $\alpha \in \mathcal{A}$ in $\mathcal{Q}$%and $\sigma_{\QBAF'}(\alpha)$ denote the strength of $\alpha \in \mathcal{A}'$ in $\mathcal{Q}'$
. 
% and $\sigma_{\QBAF'}(\alpha)$ denote the strength of $\alpha \in \mathcal{A}'$ in $\mathcal{Q}'$, respectively.
%\end{notation}

\section{Argumentative LLMs} 
\label{sec:main}

%In this section, we introduce our 
Our method is based on extracting interpretable %argumentative frameworks from the outputs of
QBAFs from LLMs and formally reasoning with them with gradual semantics. As shown in Figure \ref{fig:pipeline}, there are three integral components in ArgLLMs: argument generation, intrinsic argument strength attribution and argumentative strength calculation. %We retain any notation previously used in Section \ref{sec:preliminaries}.
We outline them here, for any input claim $x$ (see \S\ref{sec:preliminaries}),
and %provide 
give implementation details for claim verification in \S\ref{sec:experiments}.

\paragraph{Argument generation}
%{\bf Argument generation} 
Argument generation returns a BAF $\mathcal{B}$ for $x$ (see \S\ref{sec:preliminaries}):
\begin{equation*}
\Gamma(x, G, \theta) \rightarrow \mathcal{B}.
\end{equation*}
Here, $\Gamma$ %represents the
is a \emph{generating function}%\todo{not a function really...method? process?} 
%(LLM prompting in %\S\ref{sec:experiments}
%this paper)
, %$x$ is the input claim, 
$G$ denotes the underpinning \emph{generative model}%(LLM in %\S\ref{sec:experiments} this paper)
, and $\theta$ represents the \emph{parameters} associated with the argument generation, e.g. the split between attackers and supporters for each argument and the desired \emph{width} and \emph{depth} of the BAF, determined respectively by the number and length of paths from $x$ to leaves in $\mathcal{B}$. 
%The output $B$ is a BAF (see Section \ref{sec:preliminaries} for details).
For illustration, in Figure~\ref{fig:example}, we set $\theta$ to generate a BAF with width 2 and depth 1. Note that, in general, $\theta$ may influence $G$ to determine 
depth and width dynamically, e.g. based on the semantic %novelty
originality of the incrementally generated arguments. 

Previous work has %demonstrated
shown that LLMs are able to effectively generate counter-arguments given a preexisting argument \cite{Chen2023ExploringTP,Furman_23}. Leveraging this capability, we use LLMs (in conjunction with prompts) as the underpinning generative model $G$, to %perform an extension of this task, acting (in conjunction with prompts) as the generative model, $G$, to 
produce both %the 
attacking and supporting arguments %required 
to populate a BAF. 
%In our case the root argument, $x$, is a (domain-agnostic) claim. We derive these claims from existing QA datasets (for an example, see Figure \ref{fig:example}). In our case we experiment with a range of parameters, $\theta$, specified in Section \ref{sec:experiments}.

\paragraph{Intrinsic argument strength attribution}
%{\bf Intrinsic argument strength attribution} 
Intrinsic argument strength attribution adds base scores to the BAF, to obtain a QBAF $\QBAF$ for $x$:
\begin{equation*}
    \mathcal{E}(\BAF, E) \rightarrow \QBAF.
\end{equation*}
Here, %$\epsilon$ as 
$\mathcal{E}$ is an \emph{evaluative function} %\todo{not a function really...method? process?} 
%(based on LLM prompting %for uncertainty measurement 
%in this paper), %taking a BAF, $B$, and an evaluative model, $E$, as input and returning a QBAF, $Q$.
and $E$ is an \emph{evaluative model}% (LLM in this paper)
.
For illustration, in Figure~\ref{fig:example}, we obtain values of 0.85 for the supporter and 0.70 for the attacker (reported as confidence values in the visualisation shown).

There have been a number of previous attempts to assess the quality %, or intrinsic strength,
of arguments, which can be seen as a proxy for their intrinsic strength. These attempts have either used pairwise comparison between arguments \cite{Habernal2016WhichAI, Simpson2018FindingCA}, or human-annotated arguments \cite{lauscher-etal-2020-rhetoric}% to produce argument quality scores
.
However, producing such data is %highly 
resource intensive. Instead, in all instantiations of ArgLLMs in this paper, we rely on the knowledge embedded into the LLMs, using them as the evaluative model, $E$, zero-shot and without any task-specific fine-tuning. There have been some analogous uses of LLMs, such as for forecasting \cite{halawi2024approaching}, where the models are used to assign numerical confidences to their outputs (within the same context window).
Incidentally, our present study can be seen as assessing if this is an `emergent' capability of current LLMs to assess argument quality, as 
we deem unlikely that either the pre-training or the supervised training stages contained many instances of this fairly niche task. Assigning an intrinsic strength to an argument is quite subjective, and so a direct comparison between human and machine ratings may not be an ideal analysis (as it is highly likely that there would be a large variation between human scores for an individual argument). Thus, using the scores for an objective-driven, empirical task (claim verification), and ascertaining their suitability post-hoc, is perhaps a more effective method of assessing this capacity in LLMs.

\paragraph{Argumentative strength calculation}
%{\bf Argumentative strength calculation} 
Argumentative strength calculation amounts to applying a gradual semantics to resolve the conflicts within the QBAF and obtain an assessment of $x$:
\begin{equation*}
\Sigma(x, \QBAF, \sigma) \rightarrow \sigma_{\QBAF}(x).
\end{equation*}
Here, $\Sigma$ represents the %resolution 
\emph{strength calculation function}, %where $x$ is the initial claim, $Q$ is the QBAF, and $\sigma_Q$ is the argument semantics applied to the QBAF. 
and $\sigma$ (DF-QuAD in the main experiments in \S\ref{sec:experiments}%this paper
) is a chosen gradual semantics. The output $\sigma_{\QBAF}(x)$ represents the evaluated strength (or degree of acceptability) of the %initial claim 
input $x$ according to  $\sigma_{\QBAF}$,
taking into account the structural (dialectical) and quantitative (base score) aspects of the QBAF%, producing a final evaluation of the claim’s acceptability based on the generated arguments and their attributed strengths. %DF-Quad etc.?
.
For illustration, in Figure~\ref{fig:example}, we obtain a value of 0.75 for the claim (reported as confidence value in the %visualisation shown
figure).

The strength of %the input 
$x$ %\todo{XY: change x to greek style to be consistent with the content in Sec6?} 
can be seen as the final output of ArgLLMs but it can also be used to determine further outputs, e.g., in the case of claim verification, to determine whether the input claim $x$ is true or false.
In Figure~\ref{fig:example},
given that the computed value for the claim is above  0.5, the ArgLLM returns the label True
(we use 0.5 as threshold in all our experiments in \S\ref{sec:experiments}). While we did not empirically validate or tune the value of this threshold, we chose it for suitability for the task of claim verification and ease of empirical validation.

In addition to the computed value/decision, ArgLLMs also return the computed QBAF as the reasoning trail for the outputs %. In this sense, ArgLLMs 
and thus can be deemed to be interpretable.  
% ...Furthermore, the composite nature of the reasoning results in highly explainable and contestable outputs. 
The output can be directly attributed to the generated arguments and their associated strengths,  and the QBAF can serve as an explanation for why a decision has been made. 
In addition, should the QBAF be large, 
explanations in various forms can be extracted from the QBAFs, e.g. 
argument attributions~\cite{Yin2023ArgumentAE,KampikPYCT24},
% \todo{fix reference, ECAI only+ add the JAR paper with Tim}
relation attributions~\cite{YIN_RAE_IJCAI}
and counterfactual% explanations
s~\cite{YIN_CEQ}.
% facilitates the realisation of integrating our methodology into such a human-computer hybrid system. For large argument graphs, one is able to automatically surface the arguments which have the greatest impact on the final score. 
These %forms  of 
explanation would allow human users to manually check only the most significant components of the QBAFs, and their respective scores. This makes ArgLLMs amenable to human oversight, even in cases where there may be %hundreds or thousands of relevant 
very many arguments.

\section{Performance Evaluation}
\label{sec:experiments}
In this section, we describe the experiments evaluating the reasoning performance of ArgLLMs on our selected task of claim verification.\footnote{All our experiments are executed with two RTX 4090 24GB GPUs on an Intel(R) Xeon(R) w5-2455X.}
We derive claims from existing QA datasets (for an example, see Figure \ref{fig:example}). 
Also, we experiment with a range of parameters, $\theta$, in the argument generation component and with several choices for $E$. We report results for  DF-QuAD as $\sigma$ in the %argumentative strength calculation 
$\Sigma$ component, but include results for another choice of gradual semantics (quadratic energy~\cite{Potyka_18}) in the Supplementary Material in the extended version (referred to in the remainder in short as SM).

\subsection{Experimental Set-up}

\paragraph{Datasets}
%\label{sec:dataset}
We focus %focused 
on three %different
claim verification datasets adapted from existing Q/A datasets: TruthfulClaim (adapted from TruthfulQA~\cite{truthfulqa21}), StrategyClaim (adapted from StrategyQA~\cite{strategyqa21}) and MedClaim (adapted from MedQA~\cite{medqa20})\iffalse \footnote{\note{Our %versions of the 
datasets are publicly available at \url{https://anonymous.4open.science/r/ArgumentativeLargeLanguageModelsforExplainableandContestableDecisionMaking-93BE/README.md}}}
\fi
. For the purposes of our experiments, we transformed the question-answer pairs in each of the original datasets into claims with true/false labels. We did not use the original data directly as the LLMs we experimented with did not perform adequately when evaluating the validity of question-answer pairs rather than self-contained claims. To generate the claims for each Q/A pair, we used LLMs, manually checking and amending the results to ensure their correctness.
The three datasets have different flavours, reflecting the original datasets': 
%The TruthfulClaim dataset is based on TruthfulQA~\cite{truthfulqa21} and 
TruthfulQA was curated specifically to evaluate if LLMs are able to identify truthful answers without being deceived by common misconceptions and falsehoods; %StrategyClaim, adapted from StrategyQA~\cite{strategyqa21}, is 
StrategyQA was designed to evaluate whether LLMs can strategically reason; %. Finally, the MedClaim dataset adapted from MedQA~\cite{medqa20} evaluates
and MedQA evaluates models on claims associated with medical problems from the professional medical board exams. 
%In contrast with the other two source datasets, the MedQA dataset additionally contains contextual information for each question. This contextual information is considered by the model along with each claim, and the task on this dataset thus becomes \emph{conditioned claim verification}.
TruthfulClaim and StrategyClaim embed unconditioned claim verification, whereas  MedClaim
embeds conditioned claim verification.
For our experiments, we %selected 
select 700 claims from TruthfulClaim and  StrategyClaim (200 for the %initial
prompt selection experiments, as discussed later, and 500 for the main experiments), and 500 claims from the MedClaim dataset for the main experiments. All the datasets we use for our main experiments are balanced (i.e. 250 True and 250 False labels). The %reason for selecting a 
restriction to subsets of the datasets is due to the resource cost %associated with experimenting 
with LLMs% on bigger datasets
.

\paragraph{LLMs}
%\label{sec:llms}
We %consider four models in our experiments
use seven main models: Mistral (Mistral-7B-Instruct-v0.2)~\cite{jiang2023mistral}, Mixtral (Mixtral-8x7B-Instruct-v0.1)~\cite{jiang2024mixtral}, Gemma (gemma-7b-it)~\cite{gemmateam2024gemma}, Gemma 2 (gemma-2-9b-it)~\cite{gemma2}, Llama 3 (Meta-Llama-3-8B-Instruct)~\cite{llama3}, GPT-3.5-turbo (GPT-3.5-turbo-0125) \cite{Brown2020LanguageMA} and GPT-4o mini (gpt-4o-mini)~\cite{gpt-4o-mini}. We chose Mistral, Mixtral, Gemma and Llama 3 as they were among the most known and best-performing open-source\footnote{We consider a broad notion of the term ``open-source", not necessarily implying the use of OSI-approved licenses.} models of reasonable size at the time of our evaluation. In order to reduce the computational costs of running the open-source models, we quantise them to 4 bits \cite{dettmers2024qlora} %when running our experiments 
for both the baselines and our method. As representatives of models with proprietary weights, we chose GPT-3.5-turbo and GPT-4o mini, since they had the best performance/cost trade-off. For all the models, we %used
use parameters:  temperature 0.7, max new tokens for arguments 128, max new tokens for baselines 768, top-p 0.95 and repetition penalty 1.0.

% We did not use Llama-2~\cite{touvron2023llama}, as its smaller variants are typically ranked worse compared to the selected models and since the Llama-2 70B model (which is the biggest Llama-2 model) did not perform well on the validation dataset.
% (using bits-and-bytes~\cite{dettmers2023qlora}).

\paragraph{Baselines} We compare our method with three baselines: direct questioning (``Direct Question" in short), estimated confidence (``Est. Confidence" in short) and questioning with chain-of-thought~\cite{chain-of-thought2022} (``Chain-of-Thought" in short). The direct questioning baseline consists of directly asking the LLMs if the given claim is true or false by prompting, where we constrain the output of the open-source %models 
LLMs to true or false. For the estimated confidence, we ask the LLMs for a numerical confidence score for the given claim, ranging from 0 to 100. Similarly to direct questioning, we constrain the output of the open-source %models 
LLMs to the associated values. We consider the claim to be predicted as truthful if the outputted confidence is greater than 50 and false otherwise. Finally, Chain-of-Thought %prompting~\cite{chain-of-thought2022} as our third baseline. This prompt-based technique 
breaks down the problem into multiple steps before the final decision is outputted. Then, we pass the reasoning back to the LLM, in a separate context window, to get the final decision.

% \subsection{Baselines}
% We compare our method with three baselines.

% \paragraph{Baseline 1: Direct questioning (Direct Question {\rm in short})}
% We directly ask the LLMs if the given claim is true or false by prompting. 
% % The prompt used for this baseline is given in Appendix~\ref{app:direct_questioning_prompt}. 
% We constrained the output of the open-source models to true/false.

% \paragraph{Baseline 2: Estimated confidence %(then true if above 50\% and else false)
% (Est. Confidence {\rm in short})}
% We ask the LLMs for a confidence score on the given claim. The confidence score ranges from 0-100. 
% % The prompt used for this baseline is given in Appendix~\ref{app:direct_questioning_confidence_prompt}.
% We constrain the output to values ranging from 0-100 for the open-source models in this baseline. Then to get a final decision (i.e. true/false) for the claim we check whether the outputted confidence is greater than 50. If it is greater than 50 then the claim is considered true, otherwise the claim is deemed false.

% \paragraph{Baseline 3: Questioning with chain-of-thought (Chain-of-Thought {\rm in short})}
% As our third baseline, we use (two-stage) chain-of-thought prompting~\cite{chain-of-thought2022}. This prompt-based technique breaks down the problem into discrete steps before the final decision is outputted. Then, we pass the reasoning step back to the LLM, in a separate context window, to get the final decision. 
% % Both prompts are given in Appendix~\ref{app:chain_of_thought_prompt}.

\paragraph{Prompt Selection}
We take a principled approach for prompt selection, as %it has been shown that the result of 
slight variations in prompting on downstream task performance can be significant \cite{santu2023teler}. To reduce the impact of prompt choice on our final evaluation, we independently devised three different prompts for both ArgLLMs and the baselines.
% (see Appendix~\ref{app:prompts} for details).

We evaluate all prompts, and the combinations thereof, in a pilot experiment on two validation sets of 200 samples each, taken from TruthfulClaim and StrategyClaim, using Mistral and Mixtral models (selected as two substantially different open-source %model architectures
models). 
In this evaluation, we separately considered the prompts for the baselines %(direct inference, estimated confidence and chain-of-thought) 
%as well as 
and  ArgLLM components% of the argumentative approach (argument generation and argument strength attribution)
.
We find a large variation in performance for any %particular 
prompt with any given dataset and model combination, both for ArgLLMs and the baselines. We choose the highest average scoring prompts over all tested models and datasets. The results of %our prompt 
these experiments are %included 
in the SM.
% (as shown in Tables \ref{tab:base_prompt_result_avg} and \ref{tab:arg_prompt_result_avg}
% % \dg{DG: There are 14 tables, better to point to appendix only.} 
% in Appendix \ref{app:prompt_exps}).

% We chose one of GPT-3.5 (GPT-4) too expensive. Chose Mistral, Mixtral and Gemma (?), as these were the best-performing publicly available models of reasonable size. We did not use LLama-2, as its smaller variants are typically ranked worse compared to the selected models and since it did not perform well during the initial experiments.
 
% which we chose, why, why we did not choose others...

\begin{figure}[]
    \centering
    \includegraphics[width=0.35\textwidth]{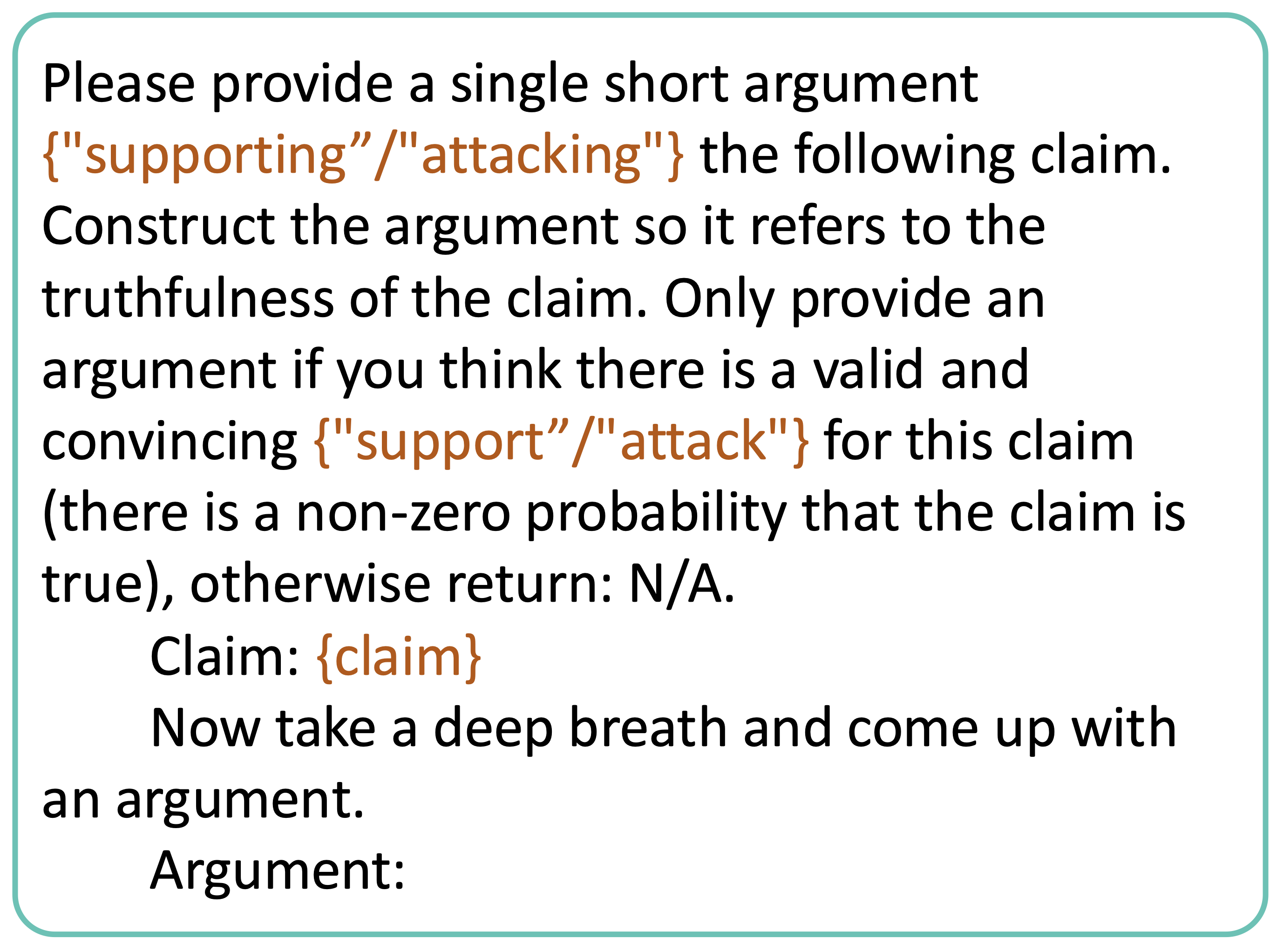}
    \caption{Prompt used for %argument generation. 
    $\Gamma$. \{``supporting''/``attacking''\} and \{``support''/``attack''\} are %conditional to the required argument type
    determined by $\theta$.}
    % (i.e. if a support argument is required, the conditionals would be "supporting" and "support", respectively). In our prompt, \{claim\} is replaced with the claim we want to verify.}
    \label{fig:main_am_prompt}
\end{figure}

\begin{figure}[ht!]
    \centering
    \includegraphics[width=0.35\textwidth]{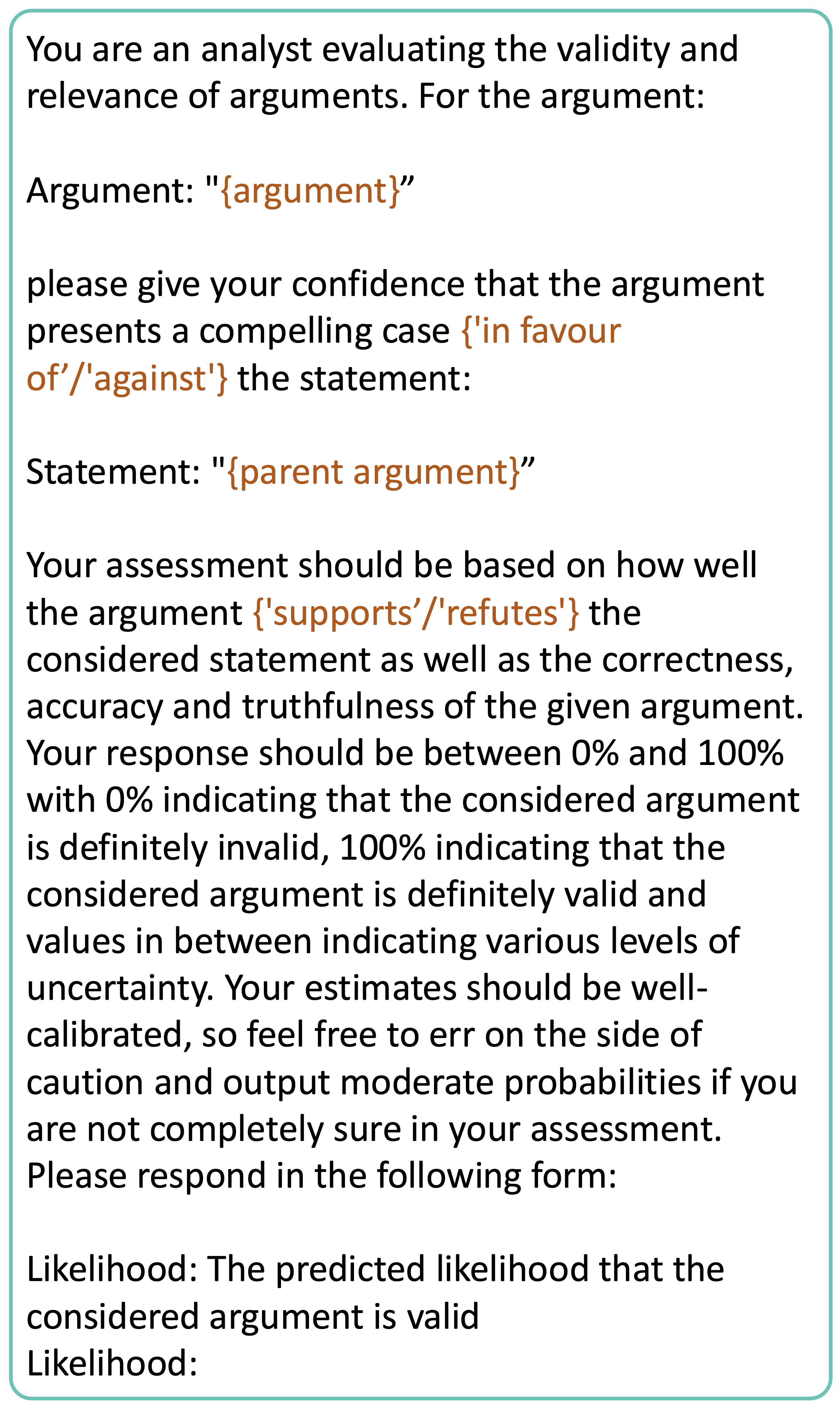}
    \caption{Prompt used for %argument strength attribution
    $\mathcal{E}$. \{``in favour of''/``against''\} and \{``supports''/``refutes''\} %are conditional to %what type of argument is given
    depend on 
    the type of \{argument\}.}
    % (i.e. if an attack argument is given the conditionals would be "against" and "refutes", respectively). In our prompt, \{argument\} is replaced with the argument that needs to be evaluated and \{parent argument\} is replaced with the parent argument of that argument.}
    \label{fig:main_ue_prompt}
\end{figure}

\paragraph{ArgLLM Variations}
In our experiments, we use four different variations of ArgLLMs. For all the variations, we use the same prompts for argument generation ($\Gamma$), given in Figure~\ref{fig:main_am_prompt}, and intrinsic argument strength attribution ($\mathcal{E}$), given in Figure~\ref{fig:main_ue_prompt}. 
%The prompt for argument generation  is given in Figure~\ref{fig:main_am_prompt} and the one for argument strength attribution ($\epsilon$) in Figure~\ref{fig:main_ue_prompt}. 
%We determine the final strength of each argument using DF-QuAD \cite{Rago_16} as the argumentation semantics (\delete{$\Sigma$}\AR{$\sigma$}). 
In order to assess accuracy for claim verification we use a similar threshold as for the Est. Confidence baseline — if the the input claim's 
%is evaluated to have a final strength of 
final strength is 
greater than 0.5 it is classified as true, and otherwise as false.

% The prompt for argument strength attribution is given in Figure~\ref{fig:main_ue_prompt}.

% \note{could go in the caption so it takes less space} In the prompt used for argument strength attribution \{"in favour of"/"against"\} and \{"supports"/"refutes"\} are conditional to what type of argument is given (i.e. if an attack argument is given the conditionals would be "against" and "refutes", respectively). In our prompt, \{argument\} is replaced with the argument that needs to be evaluated and \{parent argument\} is replaced with the parent argument of the argument that needs to be evaluated.

The four variations we consider arise from %two different 
the hyperparameters $\theta$ and
the choice of intrinsic strength for the input claim by $\mathcal{E}$. 
%the choice of the base score for the evaluated claim and the depth of the generated argumentation frameworks.
For $\theta$, we consider two options — Depth=1 and Depth=2: BAFs with Depth=1 are composed of the claim along with two generated arguments (a supporter and an attacker); in BAFs with Depth=2, we recursively generate a supporter and an attacker for each of the arguments in Depth=1, giving seven arguments in total. 
The input claim can either be given a neutral score of $0.5$ (0.5 Base Arg in short), which means that its final strength will be solely determined by the remaining arguments in the QBAF, or can be assigned a confidence score %in a way analogical to 
as for the other arguments (Est. Base Arg in short). 
By considering the combinations of the above %hyperparameter 
choices, we get the $2 \times 2 = 4$ variations.

\subsection{Evaluation Results}
\label{sec:results}

\begin{table*}[t]
\centering

\small
\begin{tabular}{ccccc|cccc}
\toprule
&  
& \textbf{\makecell{Direct\\Question}}  
& \textbf{\makecell{Est. \\ Confidence}}  
& \textbf{\makecell{Chain-of-\\Thought}}  
& \textbf{\makecell{0.5 Base\\Arg (D=1)}}  
& \textbf{\makecell{0.5 Base\\Arg (D=2)}}  
& \textbf{\makecell{Est. Base\\Arg (D=1)}}  
& \textbf{\makecell{Est. Base\\Arg (D=2)}}  \\ \toprule

\textbf{\multirow{7}{*}{\makecell{Truthful\\Claim}}} 
& Mistral  & \underline{0.73} & \underline{0.73} & \underline{0.75} & 0.65 & 0.67 & \textbf{0.76} & \underline{0.75} \\ 
& Mixtral  & 0.77 & 0.77 & 0.76 & 0.72 & 0.72 & \textbf{0.81} & \underline{0.78} \\ 
& Gemma 7B  & \underline{0.65} & 0.62 & \textbf{0.68} & 0.64 & 0.62 & 0.63 & 0.62 \\ 
& Gemma 2 9B & \textbf{0.78} & 0.74 & \textbf{0.78} & 0.68 & 0.68 & 0.73 & 0.73 \\
& Llama 3 8B & 0.66 & \underline{0.69} & \textbf{0.70} & 0.63 & 0.61 & \underline{0.68} & \underline{0.69} \\
& GPT-3.5-turbo & 0.70 & \underline{0.73} & \textbf{0.74} & 0.60 & 0.64 & \underline{0.73} & \underline{0.72} \\ 
& GPT-4o mini & \underline{0.78} & \underline{0.79} & \underline{0.79} & 0.74 & \underline{0.78} & \textbf{0.81} & \textbf{0.81} \\ \hline

\textbf{\multirow{7}{*}{\makecell{Strategy\\Claim}}} 
& Mistral  & 0.60 & 0.61 & \textbf{0.68} & 0.58 & 0.61 & 0.62 & 0.60 \\ 
& Mixtral  & \underline{0.68} & \underline{0.67} & 0.64 & 0.62 & 0.66 & \underline{0.68} & \textbf{0.70} \\ 
& Gemma 7B  & 0.55 & \underline{0.56} & \underline{0.58} & \underline{0.56} & \textbf{0.59} & \underline{0.57} & \underline{0.57} \\ 
& Gemma 2 9B & \underline{0.71} & \underline{0.71} & \textbf{0.72} & 0.65 & 0.67 & \underline{0.71} & \underline{0.70} \\
& Llama 3 8B & 0.61 & 0.59 & \textbf{0.65} & 0.54 & 0.53 & 0.61 & 0.58 \\
& GPT-3.5-turbo & \textbf{0.73} & \underline{0.70} & \underline{0.72} & 0.56 & 0.57 & \underline{0.70} &  0.67\\
& GPT-4o mini & \textbf{0.77} & \underline{0.74} & \underline{0.75} & 0.66 & 0.67 & \underline{0.74} & \underline{0.75} \\ \hline

\textbf{\multirow{8}{*}{\makecell{Med\\Claim}}} 
& Mistral  & 0.55 & 0.57 & \textbf{0.61} & 0.50 & 0.51 & 0.53 & 0.52 \\ 
& Mixtral  & \underline{0.60} & \underline{0.62} & \underline{0.61} & 0.59 & 0.56 & \underline{0.61} & \textbf{0.63} \\ 
& Gemma 7B  & \underline{0.52} & \underline{0.53} & \textbf{0.55} & 0.51 & \underline{0.52} & \underline{0.52} & \underline{0.52} \\ 
& Gemma 2 9B & \underline{0.61} & 0.58 & \textbf{0.62} & 0.57 & \underline{0.59} & \underline{0.60} & \underline{0.59} \\
& Llama 3 8B & 0.54 & \underline{0.56} & \textbf{0.58} & 0.51 & 0.53 & 0.53 & 0.52 \\
& GPT-3.5-turbo & \textbf{0.67} & 0.57 & \textbf{0.67} & 0.56 & 0.55  & 0.57 & 0.57 \\
& GPT-4o mini & \textbf{0.74} & \underline{0.71} & \underline{0.72} & 0.62 & 0.65 & \underline{0.71} & \underline{0.71} \\
& GPT-4o & \textbf{0.85} & \underline{0.82} & \textbf{0.85} & 0.74 & 0.76 & \underline{0.83} & 0.80 \\ \hline
% & GPT-4 & \textbf{0.66} & 0.60 & \textbf{0.66} & 0.52 & 0.56 & \underline{0.64} & 0.60 \\ \hline

% \textbf{\multirow{4}{*}{\makecell{Average}}} 
% & Mistral & 0.627& 0.638&  \textbf{0.681}& 0.573& 0.571& 0.637& 0.644
% \\ 
% & Mixtral  & 0.683&  0.687& 0.670& 0.643& 0.643& 0.701& \textbf{0.705}
% \\ 
% & Gemma 7B  & 0.574& 0.571&  \textbf{0.602}& 0.570& 0.569& 0.571& 0.571
% \\
% & GPT-3.5-turbo & 0.701& 0.665& \textbf{0.709}& 0.575& 0.575& 0.666& 0.674\\ \hline
% & Averages & 0.647& 0.637&  \textbf{0.664}& 0.585& 0.586& 0.643& 0.651\\
% \bottomrule
\end{tabular}
\caption{Accuracy of three baselines and four variations of ArgLLMs (all using $\sigma$=DF-QuAD) on claim verification tasks. The best result for each model-dataset combination is indicated in bold. Values within $0.03$ of the best results are underlined.} 
\label{tab:main_acc}
\end{table*}

%We compared the various methods on our adapted datasets, using the LLMs introduced previously.
%
\iffalse
On TruthfulClaim, the estimated base score methods exhibited higher accuracy compared to other methods. Specifically, \emph{Est. Base Arg (D=1)} had the highest accuracy of 0.758 and 0.81 on Mistral and Mixtral, respectively, and \emph{Est. Base Arg (D=2)} reached the highest accuracy of 0.748 on GPT-3.5-turbo. 
However, on Gemma 7B, \emph{chain-of-thought} had the highest accuracy of 0.68. 
For StrategyClaim, \emph{Est. Base Arg (D=2)} achieved the highest accuracy of 0.692 on Mixtral, while \emph{chain-of-thought} had the highest accuracy of 0.684 and 0.58 on Mistral and Gemma 7B, respectively. \emph{Direct Question} performed the best for GPT-3.5-turbo, with an accuracy of 0.734. 
Regarding MedClaim, \emph{chain-of-thought} recorded the highest accuracy of 0.612 and 0.546 on Mistral and Gemma 7B, respectively. Meanwhile, \emph{Direct Question} had the highest accuracy of 0.67 on GPT-3.5-turbo while \emph{Est. Base} had the highest accuracy of 0.62 on Mixtral. 
\fi 
Generally,  accuracy %of all methods
(see  Table~\ref{tab:main_acc})
varied on different datasets across different LLMs. However, Chain-of-Thought, Direct Questioning, and ArgLLMs with estimated base score for the topic argument performed best.
% ArgLLMs with  Est. Base Arg and D=2  and Chain-of-Thought performed best% overall compared to others
% . 

Table~\ref{tab:main_acc} also includes results on an extra MedClaim experiment with GPT-4o (gpt-4o-2024-08-06)%~\cite{gpt-4-tech-report}
, testing the performance of ArgLLMs when used with a larger model. The results indicate that Direct Question and Chain-of-Thought had the best accuracy, followed by Est. Base Arg (D=1). This result might indicate that even powerful models have relatively limited zero-shot capabilities in argument generation and estimation, highlighting the potential for task-specific finetuning.

% \note{Table~\ref{tab:main_acc} also includes results on an extra experiment with GPT-4 (GPT-4-0613)~\cite{gpt-4-tech-report} to test our hypothesis that both %argument generation and intrinsic argument strength attribution 
% $\Gamma$ and $\mathcal{E}$ were ineffective for smaller models, on the conditioned claim verification task % (rather than standard claim verification)
% %. Since MedClaim was unique in this sense we only carried out the extra experiment for this dataset, and 
% %encompassed by 
% with MedClaim. We used only 50 samples therefrom, due to financial constraints. The results indicated that Est. Base Arg (D=2) had the best accuracy% of 0.68
% , followed by Direct Question and Chain-of-Thought, both of which achieved an accuracy of 0.66. The stronger performance of ArgLLMs using larger models (Mixtral and GPT-4) supports this hypothesis.}

Overall, as the table shows, ArgLLMs perform comparably with Chain-of-Thought.
%, while also leading to faithful explanations (as discussed already in \S\ref{sec:main}) and contestability (see \S\ref{sec:contestability}). 
However, perhaps the most important features of %our 
ArgLLMs cannot be adequately captured by quantitative %performance 
metrics such as accuracy. One %of these features 
such feature is that the outputs generated are faithful explanations in terms of arguments, from which decisions can be deterministically derived, rather than decisions %directly
alone. While Chain-of-Thought can also be seen as providing reasons%for the output
, these %are outputted monolithically and 
may not be faithful to the true, stochastic reasoning process of the model (as %previously 
noted by \citet{turpin2024language}). In contrast, the final decision outputted by %our system 
ArgLLMs is faithfully determined by the %constituent 
arguments and the %used 
argumentation semantics.

\section{Contestability}
\label{sec:contestability}

\begin{figure*}[htb!]
    \centering
    \includegraphics[scale=0.38]{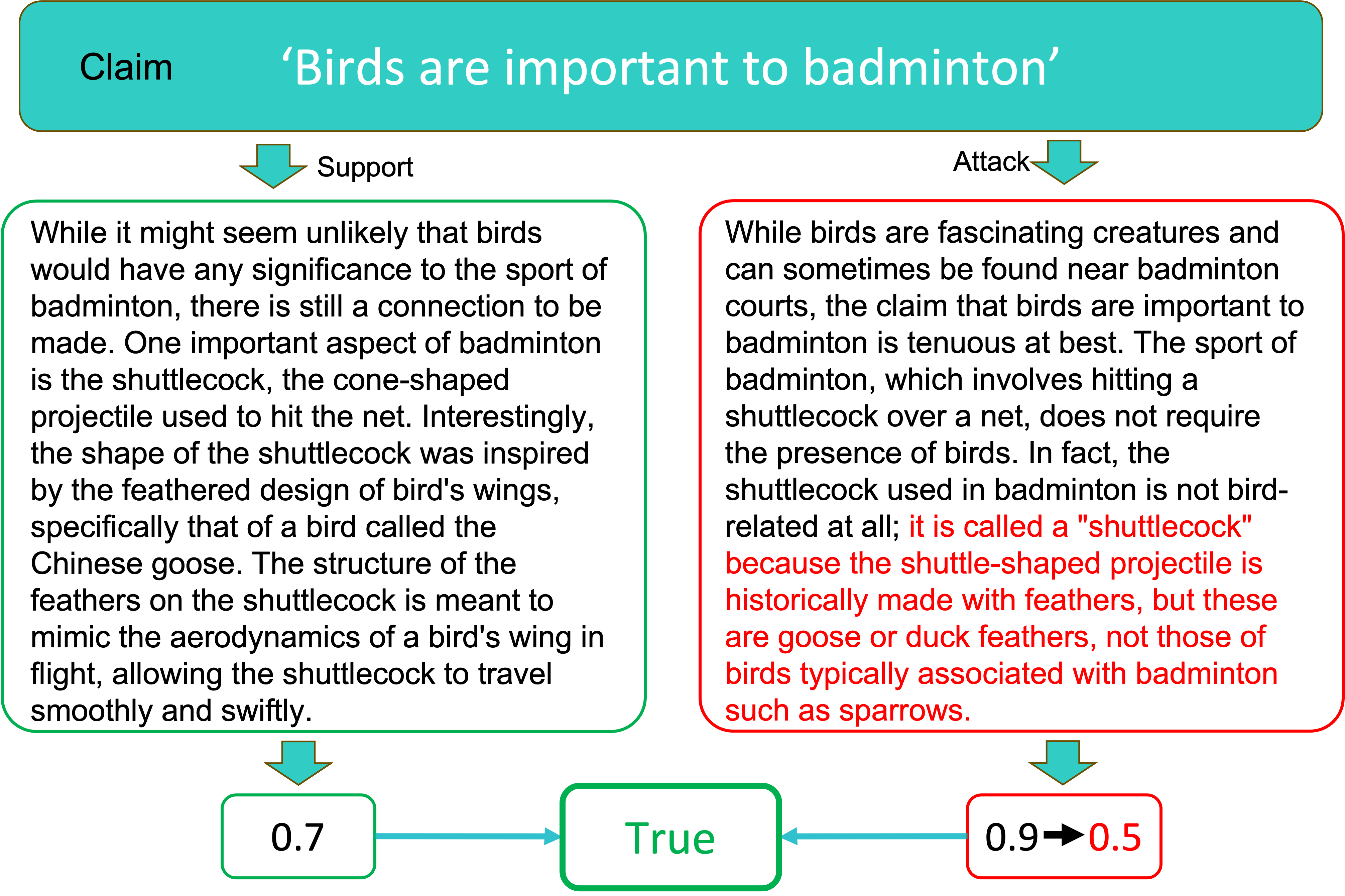}
    \caption{An example of %a user contesting 
    contestation, in the ArgLLM with Mixtral, %the intrinsic strength of an argument attacking 
    for a claim taken from StrategyClaim% (arguments are generated by Mixtral)
    . Before contestation, the claim was (incorrectly) classified as False, but after contesting the intrinsic strength of the attacking argument from 0.9 to 0.5 (citing the fallacious reasoning highlighted in red), the correct True classification results.
    % as the attacking argument was assigned a strength of 0.9, whereas the supporting argument had a strength of 0.7. However, a human user is able to modify this score (e.g. from 0.9 to 0.5), citing the fallacious reasoning present in the attacking argument (highlighted in red). This ultimately results in the correct classification.
    % note{AR: chain-of-thought also incorrect?}
    }
    \label{fig:base_alter}
\end{figure*}

Another unique feature of ArgLLMs is contestability.
The (faithful) explainability of ArgLLMs offers %users of the system 
plentiful opportunity to disagree with the reasoning, either in terms of the arguments generated being relevant or true, or the intrinsic strengths that have been attributed to them being representative of the extent to which they support or attack their `parent' argument. 
For illustration, consider   Figure~\ref{fig:base_alter}.
% \note{AR: I think we'll get attacked in that we're claiming this is the main advantage of the method but we don't evaluate it. I think we should at least demonstrate the contestability with a toy example taken from the datasets?}
Here, a user presented with the output of the ArgLLM may disagree with some parts of the argument attacking the claim. This may lead the user to decrease the argument intrinsic strength and contest the False label originally determined by the ArgLLM.
%An additional 
A further illustration can be found in the SM.

Here, we frame formally the contestability functionality for ArgLLMs, by proposing two notions of contestability in the QBAF setting. This formal analysis applies to 
QBAFs of any depth and width. 
Since QBAFs are composed of arguments/relations and base scores (intrinsic argument strength), ArgLLMs  can be contested on two fronts.

\begin{figure}[h]
    \centering
    \includegraphics[width=0.5\columnwidth]{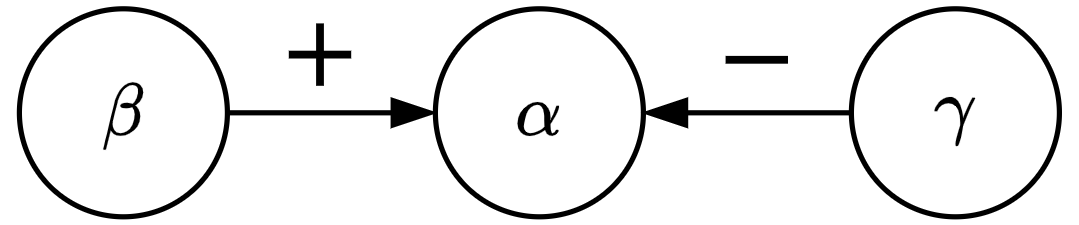}
    \caption{%Illustration of b
    Base score contestability (Example~\ref{ex:base score con}).}
    \label{fig_Q1}
\end{figure}

We start with a property governing contestability on base scores%, which states that a QBAF for an argument (in the claim verification setting, the input claim $x$) %(strength of the topic argument) 
%can be contested by increasing or decreasing the base score of any argument.
%\todo{why are these in the property environment and the properties in the background are in the definition environment? F may want to remove all environments from the background but let's see}
:

\begin{property}%[Base Score Contestability]
\label{property_basescore_contest}
%Let $M=\QBAF$ and $M'=\QBAF'$\todo{not sure this works? they're not functions of the arguments like the models are functions of the inputs. did francesca ask for this? can't we just give the relationship informally? XY: we can just delete this line},
A gradual semantics $\sigma$ satisfies \emph{base score contestability} iff for any QBAF $\QBAF$ for $\alpha^*$, for  any $\QBAF'$ with 
$\mathcal{A}' = \mathcal{A}$, 
$\mathcal{R^-}' = \mathcal{R^-}$, 
$\mathcal{R^+}' = \mathcal{R^+}$, 
and, for $\beta\in \Args$, $\tau(\beta) < \tau'(\beta)$ while $\tau(\gamma) = \tau'(\gamma)$ for all $\gamma \in \mathcal{A} \setminus \{\beta\}$:

$\bullet$  if $\beta \in \Pros(\QBAF)$, then $\sigma_{\QBAF}(\alpha^*) \leq \sigma_{\QBAF'}(\alpha^*)$;

$\bullet$ if $\beta \in \Cons(\QBAF)$, then $\sigma_{\QBAF}(\alpha^*) \geq \sigma_{\QBAF'}(\alpha^*)$.
\end{property}
%\begin{proposition}
%If $\sigma$ satisfies base score monotonicity, then $\QBAF$ satisfies base score contestability.
%\todo{this seems strange, did francesca ask for the properties to be over the QBAFs? XY: now we removed the generic def of contestability so now better?}
%\end{proposition}
This property formalises that increasing the base score of pro/con arguments in a QBAF for an argument will not decrease/increase (respectively) the argument's final strength\footnote{Pro/con arguments no longer affect $\sigma_{\QBAF}(\alpha^*)$ when it is 1 or 0.}. %\todo{ weak, can it be made any stronger? add to SM...we can point to it here...}  
\begin{example}
\label{ex:base score con}
Let $\QBAF$ be %a QBAF for $\alpha$ 
as in Figure~\ref{fig_Q1}, and $\sigma$ be DF-QuAD. Since there is only one (odd number) attack from 
$\beta$ to $\alpha$, $\beta \in \Pros(\QBAF)$. Then, increasing $\tau(\beta)$ will not decrease $\sigma_{\QBAF}(\alpha)$%, thus $\QBAF$ is base score contestable
.
\end{example}

\noindent The next property states that the output of ArgLLMs %QBAF (the strength of the topic argument) 
can be contested by adding or removing arguments and relations.

\begin{property}%[Argument Relation Contestability]
\label{property_arg_relation_contest}
% Let $M=\QBAF$ and $M'=\QBAF'$,
A gradual semantics $\sigma$ satisfies \emph{argument relation contestability} iff for any QBAF $\QBAF$ for $\alpha^*$, for any $\QBAF'$ with
$\mathcal{A}'\!=\!\mathcal{A} \!\cup \!\{\beta\} $, 
%$\mathcal{R^-} \!\subseteq \! \mathcal{R^-}'$, 
%$\mathcal{R^+} \subseteq \mathcal{R^+}'$, 
$\mathcal{R^-}' \cup \mathcal{R^+}'=\mathcal{R^-} \cup \mathcal{R^+} \cup \{(\beta,\alpha)\}$ for some $\alpha \in \mathcal{A}$,
and $\tau'(\gamma) = \tau(\gamma)$ for all $\gamma \in \mathcal{A}$:

$\bullet$  if $\beta \in \Pros(\QBAF')$, then $\sigma_{\QBAF}(\alpha^*) \leq \sigma_{\QBAF'}(\alpha^*)$;
    
    $\bullet$ if $\beta \in \Cons(\QBAF')$, then $\sigma_{\QBAF}(\alpha^*) \geq \sigma_{\QBAF'}(\alpha^*)$.
\end{property}

%\begin{proposition}
%If $\sigma$ satisfies argument relation monotonicity, then $\QBAF$ satisfies argument relation contestability.
%\end{proposition}

\begin{figure}[h]
    \centering
    \includegraphics[width=0.7\columnwidth]{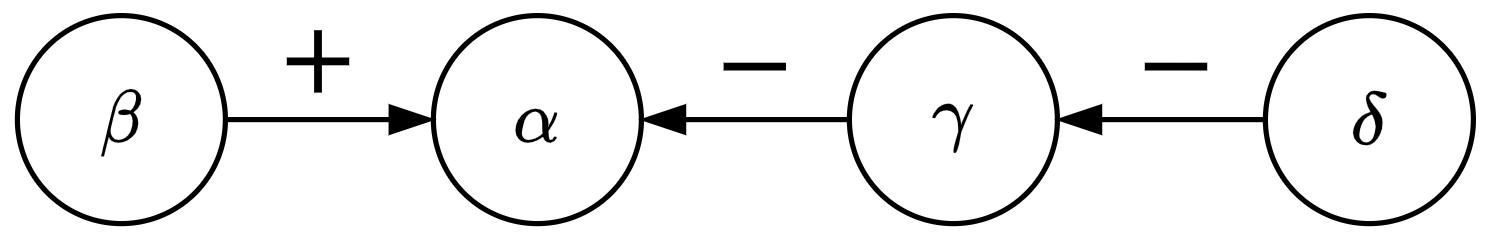}
    \caption{%Illustration of a
    Argument relation contestability (Example~\ref{ex:argcon}).
    }
    \label{fig_Q2}
\end{figure}

\begin{example}
\label{ex:argcon}
Let $\QBAF$ be as %for $\alpha$ 
in Figure~\ref{fig_Q1}, and $\sigma$ be DF-QuAD. 
If we contest $\QBAF$ by adding an attacker $\delta$ to $\gamma$ (to obtain $\QBAF'$ in Figure~\ref{fig_Q2}), then  $\delta \in \Pros(\QBAF')$ because there are two (even number) attacks from 
$\delta$ to $\alpha$. Then, $\sigma_{\QBAF}(\alpha) \leq \sigma_{\QBAF'}(\alpha)$%, so $\QBAF$ is argument relation contestable
.
\end{example}

\noindent We now give a theoretical result demonstrating DF-QuAD's suitability for providing contestability in ArgLLMs (the proof  can be found in the SM).

\begin{proposition}
DF-QuAD %and QEM satisfy 
satisfies base score contestability and argument relation contestability. %in tree QBAFs. 
\end{proposition}

%Indeed, any semantics which satisfies X (author et al., 20XX) and Y (author et al., 20XX), as DF-QuAD does, will satisfy both contestability properties (see the SM for a proof).\todo{to be added}

\noindent  Note that stronger versions of Properties~\ref{property_basescore_contest} and \ref{property_arg_relation_contest} can be defined, with strict inequalities, satisfied by the quadratic energy model gradual semantics~\cite{Potyka_18} (see the SM).

\section{Conclusions \& Future Work}
\label{sec:conclusions}

We have proposed ArgLLMs, harnessing the knowledge encoded in any LLMs to obtain interpretable outputs that can be faithfully explained and provably contested -
% general reasoning capacity of LLMs - 
without requiring any fine-tuning or external resources% - making them explainable, contestable and improving their accuracy in some circumstances
. We have shown that ArgLLMs's added value does not lead to compromising prediction accuracy  in the task of claim verification in comparison with  state-of-the-art prompting methods. 
%Furthermore, our system innately permits human-computer collaboration, and provides accurate uncertainty estimates as an output.

Our work opens several avenues for future study.
%The performance of ArgLLMs could benefit from fine-tuning for the argument generation and intrinsic argument strength attribution components. %Likewise, we expect that integrating information retrieval, both for generating the arguments and strengths, would result in further improvements.
%
%There are numerous methodologies for argument strength attribution which warrant further analysis. These include techniques which adapt our chosen method of directly prompting an LLM by, for example, 
Sampling multiple outputs of the same LLM or taking the weighted value of the relevant logits in the final layer %of the model 
could provide more sophisticated means for the intrinsic argument strength attribution component. 
Furthermore, an adapted version of the `semantic uncertainty' \cite{kuhn2022semantic} method may be devised, wherein one directly clusters semantically similar sampled arguments, rather than having to prompt models for numerical scores. 

% We also experimented with verbal confidence scores to assign argument strengths. While we did not observe promising results, this approach may respond well to supervised fine-tuning.
%

Another potential direction for the first two components, argument generation and intrinsic strength attribution, is the ensembling of many different LLMs, %both for argument generation and strength attribution
 %This is a way 
 to harness the heterogeneous knowledge encoded %in different LLMs
 therein. In a similar vein, using information retrieval or retrieval augmented generation \cite{lewis2020retrieval} may %be a way to 
 improve the %breadth and reliability of the arguments 
 generated QBAFs. We also %would also like 
 plan to conduct more %thorough 
 experiments on how tuning hyperparameters, %such as 
 e.g. temperature, would 
 impact %downstream
 performance.
 
 It would also be interesting to study %properties of 
 the generated QBAFs in terms of the properties for general argumentative explanations in the spirit of \citet{neema-argX24}, and to consider other formal properties of contestability and results for other gradual semantics for QBAFs. 
We %will 
also plan to undertake human evaluations of %the contestability and explainability of 
ArgLLMs.
Finally, 
the transparency and contestability properties of ArgLLMs make them an ideal candidate for highly complex, uncertain and high-stakes scenarios, including business, medical or legal decision-making where ArgLLMs could be used in conjunction with domain-experts %, who could review the outputted arguments and strengths, and leverage their experience and contextual knowledge to modify them accordingly. Additionally, they can add 
providing arguments of their own. 

%\clearpage
\section*{Acknowledgments}

This research was partially supported by ERC under the EU's Horizon 2020 research and innovation programme (grant agreement No. 101020934, ADIX), by J.P.
Morgan and the Royal Academy of Engineering under the Research Chairs and Senior Research Fellowships
scheme, by Imperial College through an Imperial College Research Fellowship and by UKRI through the CDT in Safe
and Trusted Artificial Intelligence (Grant No. EP/S023356/1).

\bibliographystyle{named}
\bibliography{bibliography}

\newpage

\section{Ablations: Quadratic Energy Model Semantics}
In addition to DF-QuAD, we wanted to explore the behaviour of other semantics with our method. Therefore, we have conducted ablation studies (the results are in Table~\ref{tab:ablation_acc}) using the \emph{quadratic energy model} (QEM) semantics \cite{Potyka_18}, which is defined as follows. For a given QBAF $\langle \Args, \Atts, \Supps, \BS \rangle$, the energy at $\alpha \in \Args$ is defined as $E_\alpha = \sum_{\beta \in \Supps(\alpha)}{\SF(\beta)} - \sum_{\gamma \in \Atts(\alpha)}{\SF(\gamma)}$. Then, for all $v \in \mathbb{R}$, we let $h(v) = \frac{max\{v,0\}^2}{1+max\{v,0\}^2}$. Finally, the strength of an argument $\alpha \in \Args$ is defined as $\SF(\alpha) = \BS(\alpha) + (1 - \BS(\alpha)) \cdot h(E_\alpha) -  \BS(\alpha) \cdot h(-E_\alpha)$.\footnote{We describe a simplification of the original algorithm for the case of trees, rather than (potentially cyclic) graphs.}

\begin{table*}[htp]
\centering
\caption{Accuracy of QE (left-side) and DF-QuAD (right-side) semantics of our argumentative method on claim verification tasks. The best performing method for each model-dataset combination is indicated in bold.}

\small
\begin{tabular}{cccccc|cccc}
\toprule
& & \multicolumn{4}{c}{\bf Quadratic Energy Model}
& \multicolumn{4}{c}{\bf DF-QuAD} \\
\cmidrule(lr){3-6} \cmidrule(lr){7-10}
&  
& \textbf{\makecell{0.5 Base \\ Arg (D=1)}}  
& \textbf{\makecell{0.5 Base \\ Arg (D=2)}}  
& \textbf{\makecell{Est. Base \\ Arg (D=1)}}  
& \textbf{\makecell{Est. Base \\ Arg (D=2)}}
& \textbf{\makecell{0.5 Base \\ Arg (D=1)}}  
& \textbf{\makecell{0.5 Base \\ Arg (D=2)}}  
& \textbf{\makecell{Est. Base \\ Arg (D=1)}}  
& \textbf{\makecell{Est. Base \\ Arg (D=2)}}  \\ \toprule

\textbf{\multirow{7}{*}{\makecell{Truthful\\Claim}}} 
& Mistral  & 0.65 & 0.67 & 0.75 & 0.74 & 0.65 & 0.67 & \textbf{0.76} & 0.75  \\ 
& Mixtral  & 0.72 & 0.72 & 0.80 & 0.78 & 0.72 & 0.72 & \textbf{0.81} & 0.78  \\
& Gemma 7B  & \textbf{0.64} & 0.62 & 0.63 & 0.62 & \textbf{0.64} & 0.62 & 0.63 & 0.62  \\
& Gemma 2 9B & 0.68 & 0.68 & \textbf{0.74} & \textbf{0.74} & 0.68 & 0.68 & 0.73 & 0.73  \\
& Llama 3 8B & 0.63 & 0.64 & \textbf{0.69} & 0.68 & 0.63 & 0.61 & 0.68 & \textbf{0.69}  \\
& GPT-3.5-turbo & 0.60 & 0.64 & 0.73 & 0.73 & 0.60 & 0.64 & \textbf{0.74} & 0.72  \\
& GPT-4o-mini & 0.74 & 0.78 & 0.80 & 0.80 & 0.74 & 0.78 & \textbf{0.81} & \textbf{0.81}  \\ \hline

\textbf{\multirow{7}{*}{\makecell{Strategy\\Claim}}} 
& Mistral  & 0.58 & 0.60 & \textbf{0.62} & 0.60 & 0.58 & 0.61 & \textbf{0.62} & 0.60  \\
& Mixtral  & 0.62 & 0.64 & 0.69 & 0.69 & 0.62 & 0.66 & 0.68 & \textbf{0.70}  \\
& Gemma 7B  & 0.56 & \textbf{0.59} & 0.56 & 0.56 & 0.56 & \textbf{0.59} & 0.57 & 0.57  \\ 
& Gemma 2 9B & 0.65 & 0.65 & \textbf{0.71} & 0.70 & 0.65 & 0.66 & \textbf{0.71} & 0.70  \\
& Llama 3 8B & 0.54 & 0.57 & 0.59 & 0.59 & 0.54 & 0.53 & \textbf{0.61} & 0.58  \\
& GPT-3.5-turbo & 0.56 & 0.57 & 0.70 & \textbf{0.71} & 0.56 & 0.57 & 0.69 & 0.67  \\
& GPT-4o-mini & 0.66 & 0.68 & \textbf{0.75} & \textbf{0.75} & 0.66 & 0.67 & 0.74 & \textbf{0.75}  \\ \hline

\textbf{\multirow{7}{*}{\makecell{Med\\Claim}}} 
& Mistral  & 0.50 & 0.51 & 0.55 & \textbf{0.57} & 0.50 & 0.51 & 0.53 & 0.52  \\ 
& Mixtral  & 0.59 & 0.57 & \textbf{0.63} & \textbf{0.63} & 0.59 & 0.56 & 0.61 & \textbf{0.63}  \\
& Gemma 7B  & 0.51 & \textbf{0.52} & \textbf{0.52} & \textbf{0.52} & 0.51 & \textbf{0.52} & \textbf{0.52} & \textbf{0.52}  \\
& Gemma 2 9B & 0.57 & 0.59 & 0.59 & 0.59 & 0.57 & 0.59 & \textbf{0.60} & 0.59  \\
& Llama 3 8B & 0.51 & 0.53 & \textbf{0.55} & \textbf{0.55} & 0.51 & 0.53 & 0.53 & 0.52  \\
& GPT-3.5-turbo & 0.56 & 0.55 & \textbf{0.57} & \textbf{0.57} & 0.56 & 0.55 & \textbf{0.57} & \textbf{0.57}  \\
& GPT-4o-mini & 0.62 & 0.65 & \textbf{0.71} & \textbf{0.71} & 0.62 & 0.65 & \textbf{0.71} & \textbf{0.71}  \\
& GPT-4 & 0.52 & 0.58 & 0.62 & 0.60 & 0.52 & 0.56 & \textbf{0.64} & 0.60  \\ \hline

% \textbf{\multirow{4}{*}{\makecell{Average}}} 
% & Mistral & 0.627& 0.638&  \textbf{0.681}& 0.573& 0.571& 0.637& 0.644
% \\ 
% & Mixtral  & 0.683&  0.687& 0.670& 0.643& 0.643& 0.701& \textbf{0.705}
% \\ 
% & Gemma 7B  & 0.574& 0.571&  \textbf{0.602}& 0.570& 0.569& 0.571& 0.571
% \\
% & GPT-3.5-turbo & 0.701& 0.665& \textbf{0.709}& 0.575& 0.575& 0.666& 0.674\\ \hline
% & Averages & 0.647& 0.637&  \textbf{0.664}& 0.585& 0.586& 0.643& 0.651\\
% \bottomrule
\end{tabular}
\label{tab:ablation_acc}
\end{table*}

The results of the ablation study closely matched those achieved using the DF-QuAD semantics however performed slightly worse than DF-QuAD.

\section{Baseline Prompts}
% AD: Almost directly copied from the ACL submission — will likely require format tweaking

\label{app:baseline_prompts}
We show the prompts used for the baselines.

\subsection{Direct questioning}
\label{app:direct_questioning_prompt}
\begin{figure}[htp!]
    \centering
    \includegraphics[width=\linewidth]{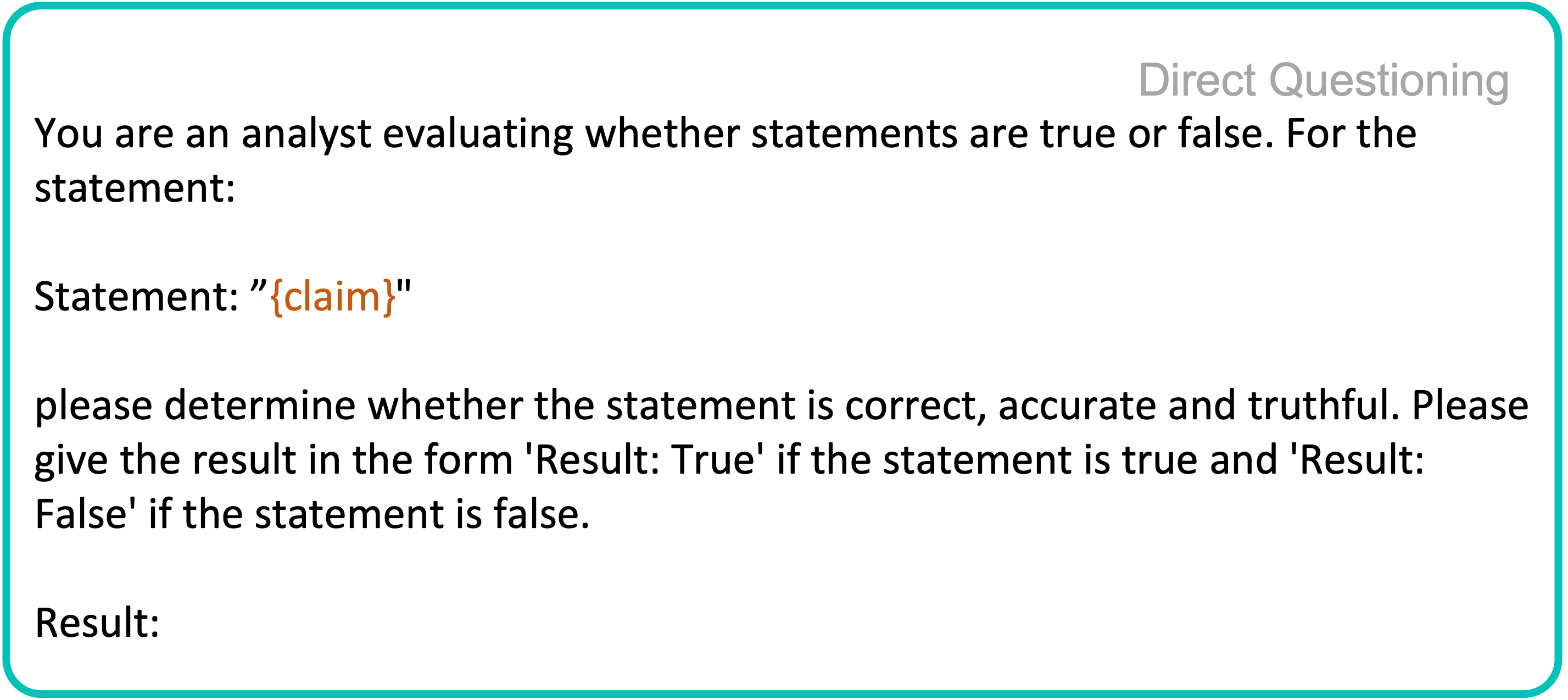}
    \caption{Prompt used for direct questioning baseline.}
    \label{fig:prompt_direct_questioning}
\end{figure}

The direct questioning prompt is shown in Figure~\ref{fig:prompt_direct_questioning}.

In our prompt, \{claim\} is replaced with the claim we want to verify.

\subsection{Direct questioning on confidence}
\label{app:direct_questioning_confidence_prompt}
\begin{figure}[ht!]
    \centering
    \includegraphics[width=\linewidth]{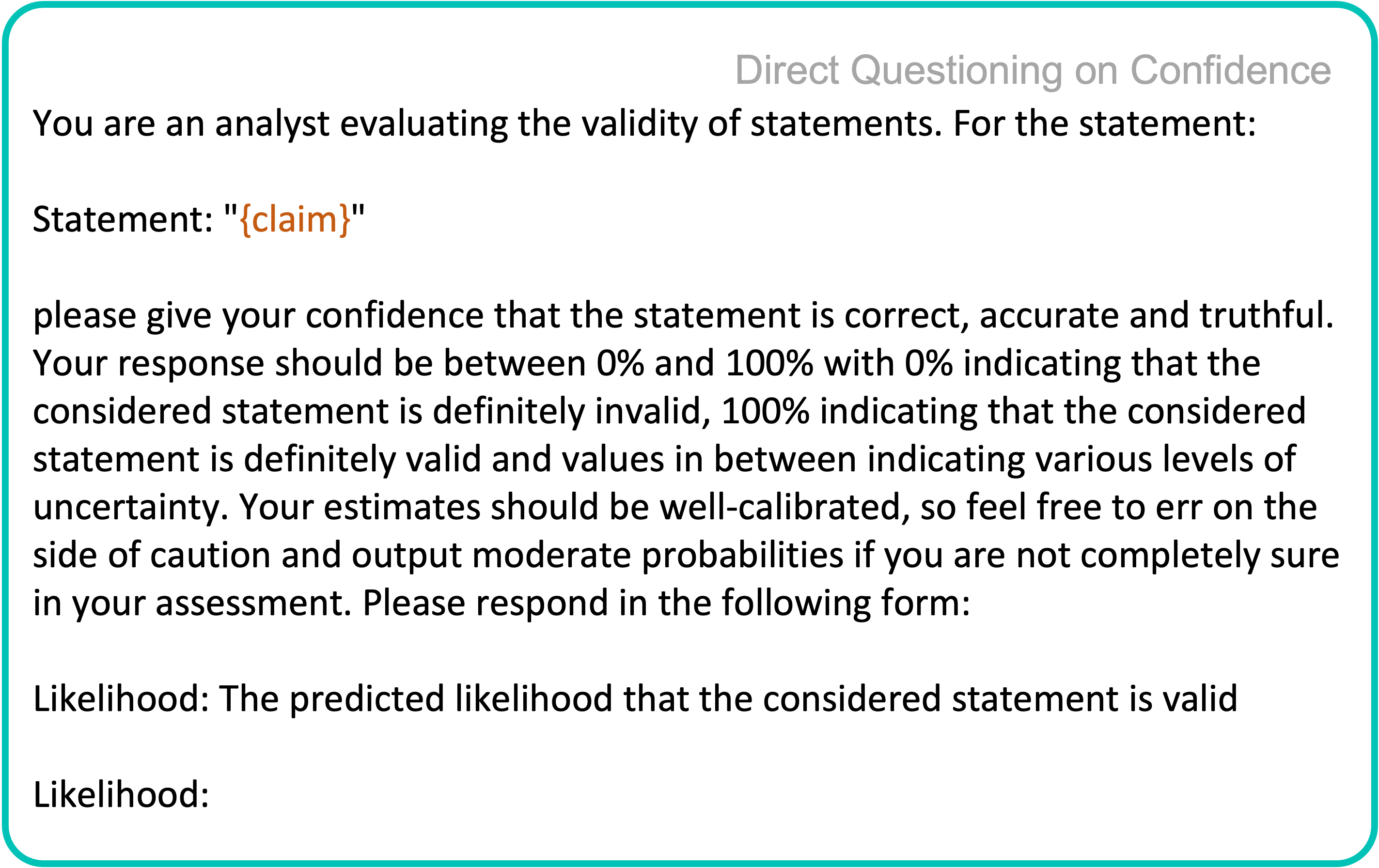}
    \caption{Prompt used for direct questioning on confidence baseline.}
    \label{fig:prompt_direct_questioning_confidence}
\end{figure}

The direct questioning on confidence prompt is shown in Figure~\ref{fig:prompt_direct_questioning_confidence}.

In our prompt, \{claim\} is replaced with the claim we want to verify.

\subsection{Questioning with Chain-of-thought}
\label{app:chain_of_thought_prompt}
\begin{figure}[ht!]
    \centering
    \includegraphics[width=\linewidth]{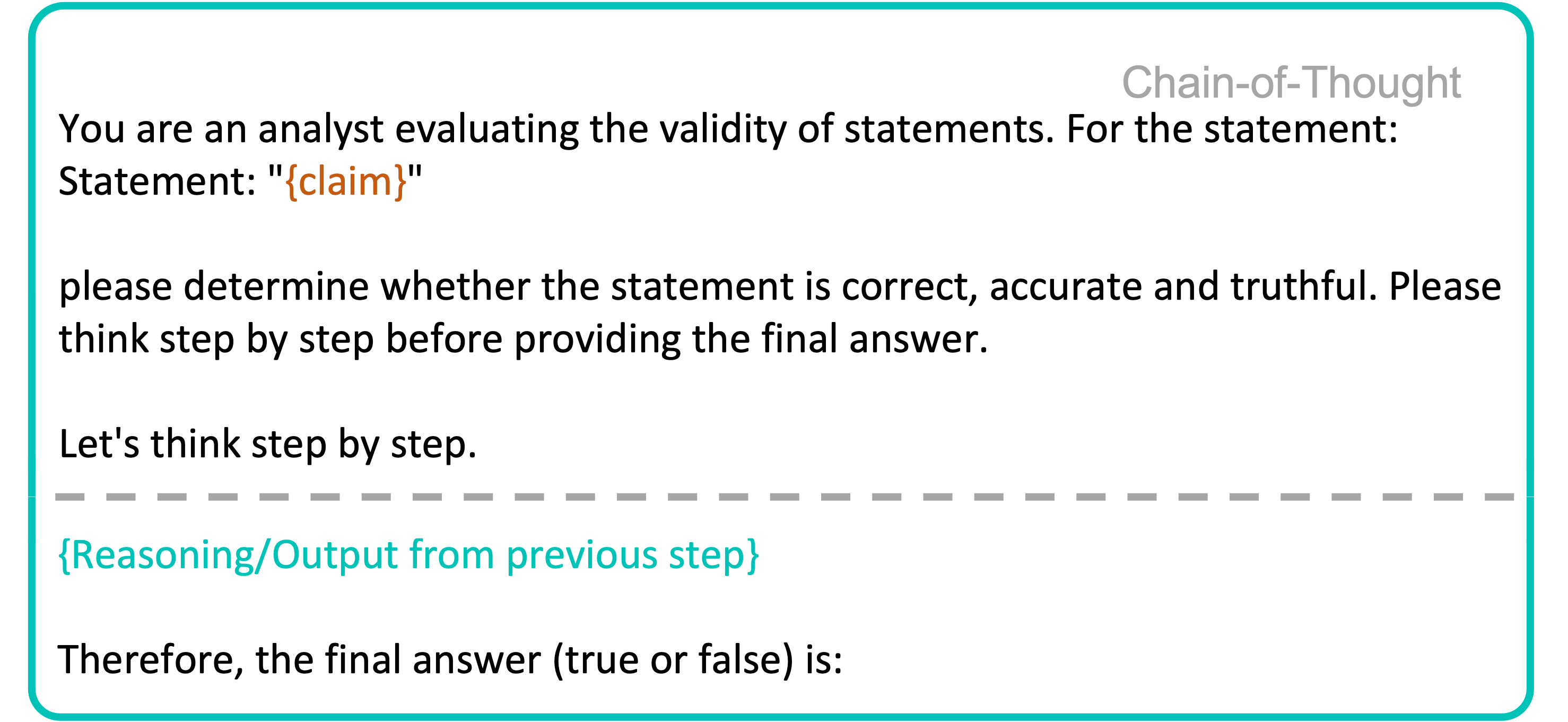}
    \caption{Prompts used for chain-of-thought baseline.}
    \label{fig:chain_of_thought_baseline}
\end{figure}

For chain-of-thought the first prompt used to obtain the discrete steps and the prompt to get the final decision are given in Figure~\ref{fig:chain_of_thought_baseline}.

The prompt above the line is to obtain the reasoning steps and \{claim\} is replaced with the claim we want to verify. The prompt below the line is to get the final decision and \{Reasoning/Output from previous step\} is replaced with the reasoning obtained from the prompt above the line.

\section{Prompt for Argument Strength Attribution for Claim}
\begin{figure}[htp!]
    \centering
    \includegraphics[width=\linewidth]{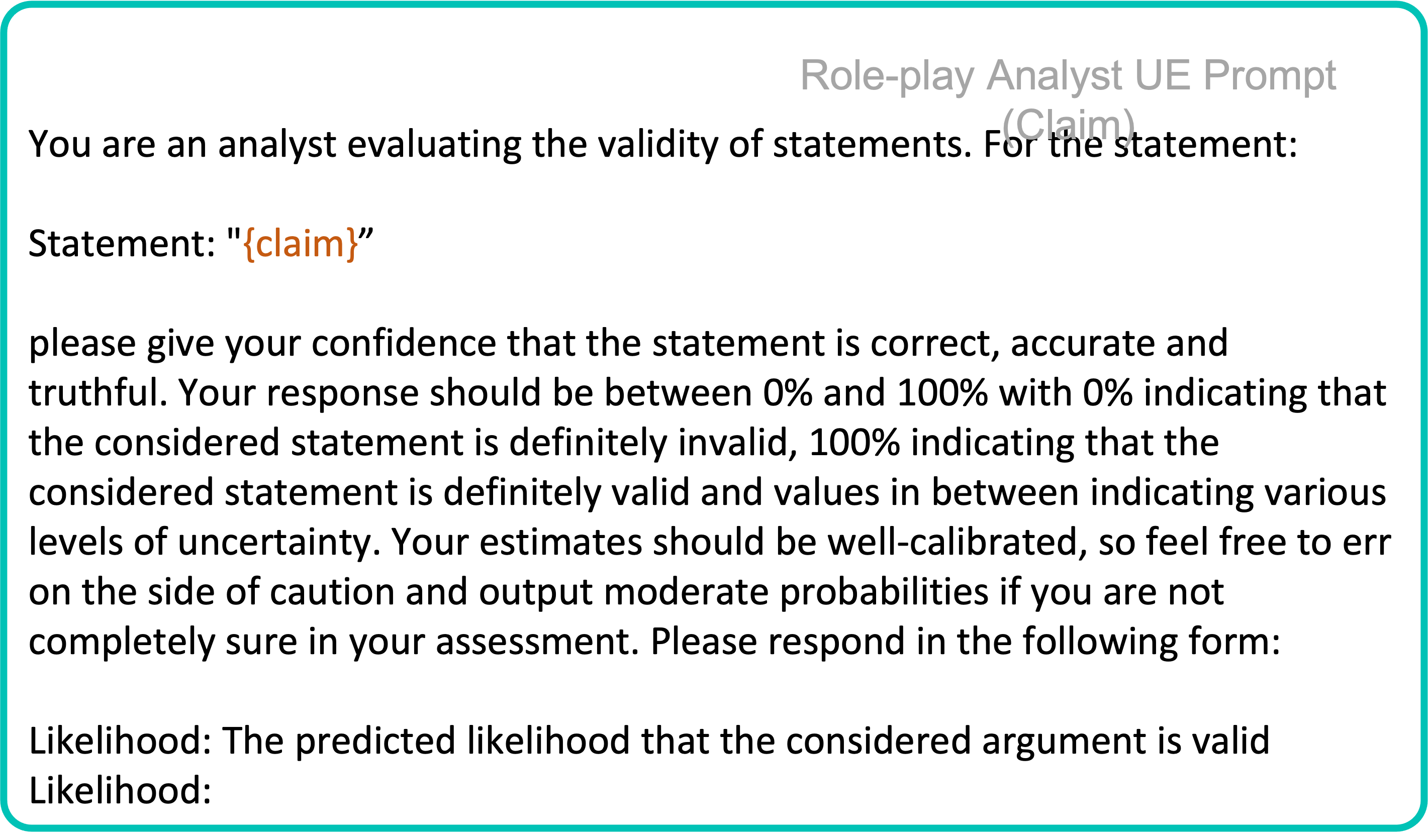}
    \caption{Prompt used for argument strength attribution for the claim. In our prompt, \{claim\} is replaced with the claim we want to verify.}
    \label{fig:main_ue_prompt_claim}
\end{figure}

The prompt for argument strength attribution does not work for the claim as it requires a parent argument to be present. So, we altered the prompt for only claim argument strength attribution (the prompt could be seen in Figure~\ref{fig:main_ue_prompt_claim}).

\section{Prompts Considered during Prompt Selection}
\label{app:prompts}

\subsection{ChatGPT prompts}
\begin{figure}[htp!]
    \centering
    \includegraphics[width=\linewidth]{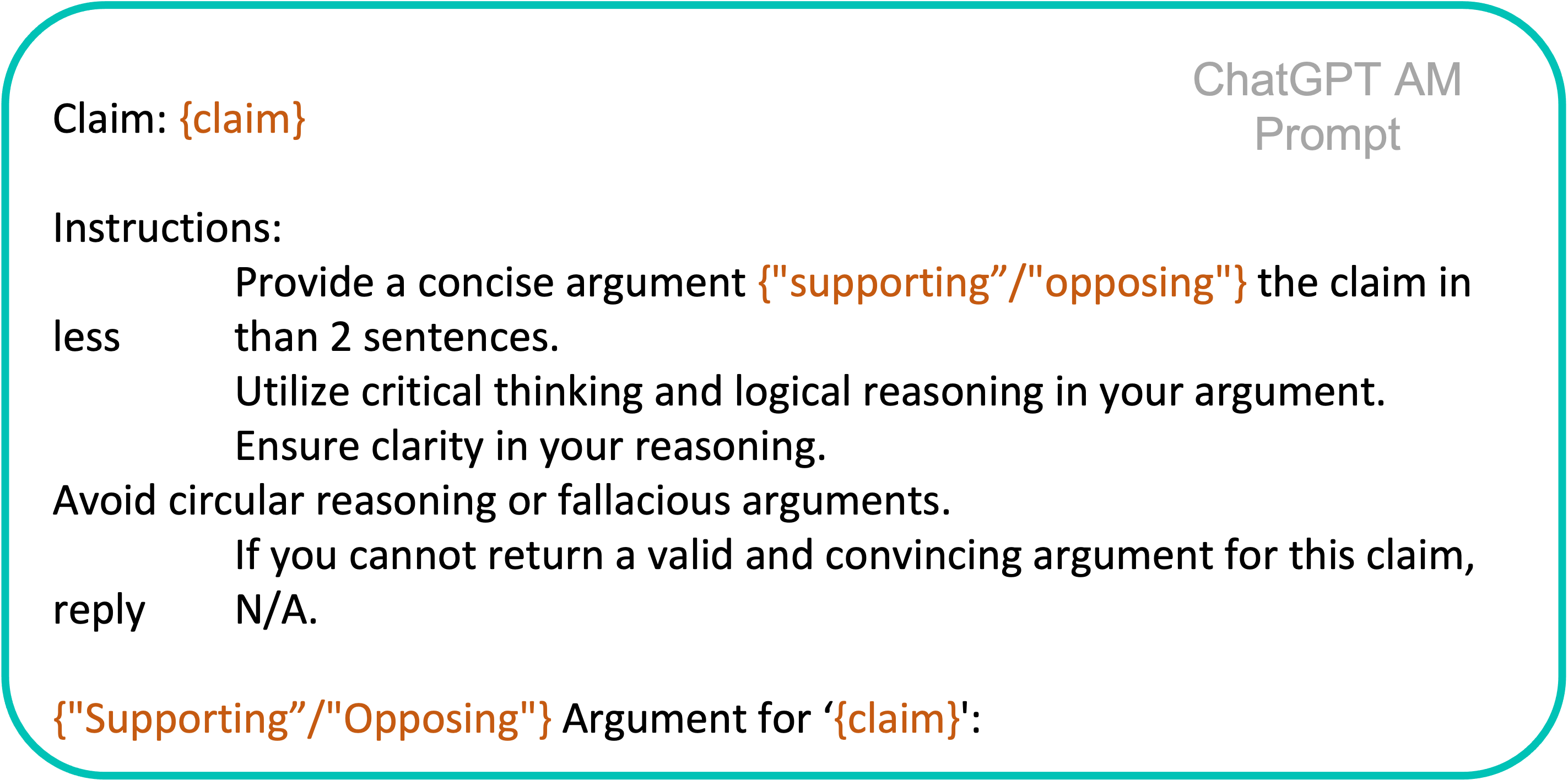}
    \caption{ChatGPT Argument Generator prompt}
    \label{fig:chatgpt_am_prompt}
\end{figure}

\begin{figure}[htp!]
    \centering
    \includegraphics[width=\linewidth]{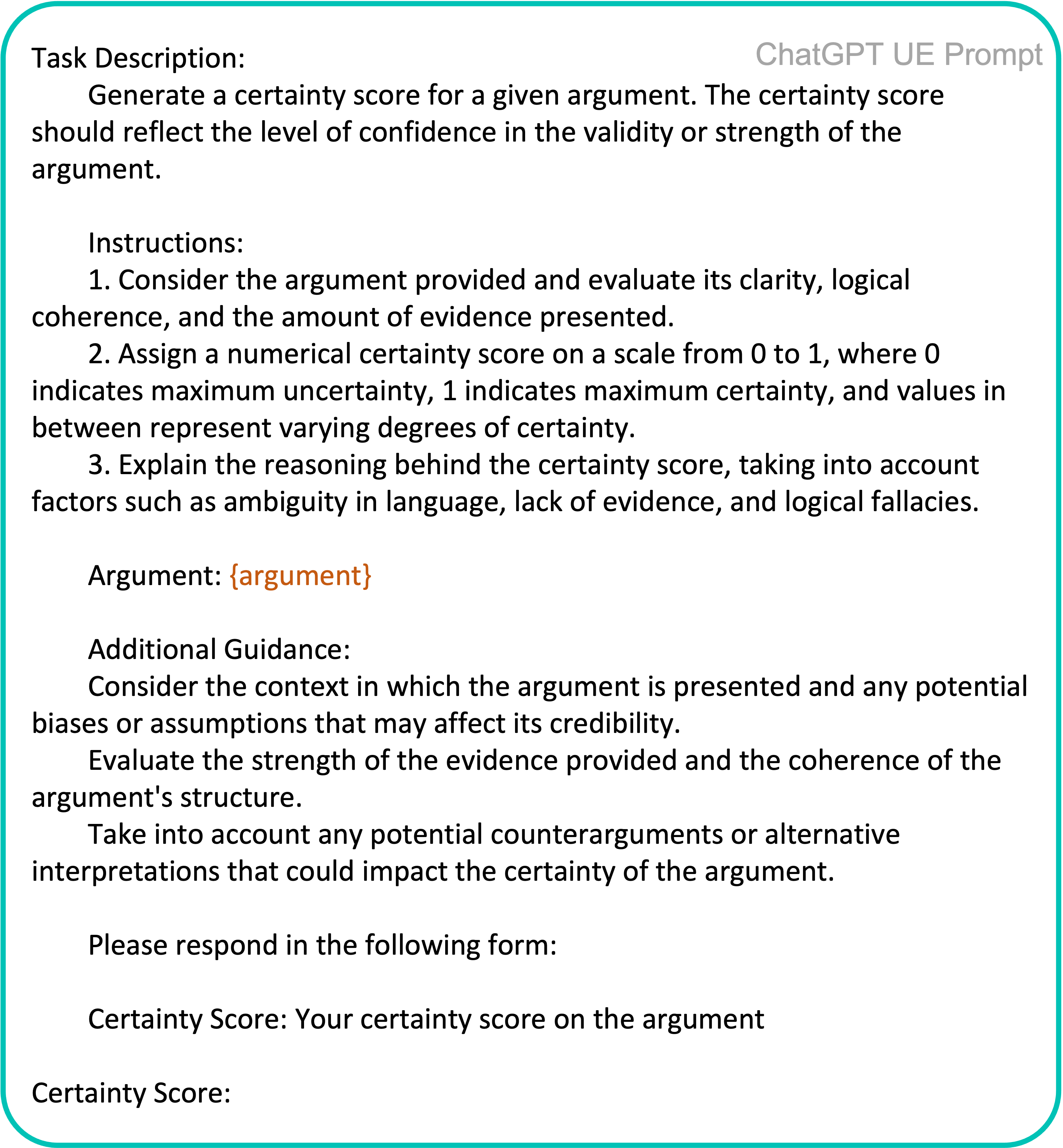}
    \caption{ChatGPT Argument Strength Attribution prompt}
    \label{fig:chatgpt_ue_prompt}
\end{figure}

ChatGPT prompts were generated mostly using ChatGPT. First, the prompt is initialised by giving ChatGPT the instructions, then the prompt is optimised by giving ChatGPT some outputs and asking it to improve the prompts. The ChatGPT Argument Generator prompt can be found in Figure~\ref{fig:chatgpt_am_prompt} and the ChatGPT Argument Strength Attribution prompt can be found in Figure~\ref{fig:chatgpt_ue_prompt}.

\subsection{Role-player prompts}
\begin{figure}[htp!]
    \centering
    \includegraphics[width=\linewidth]{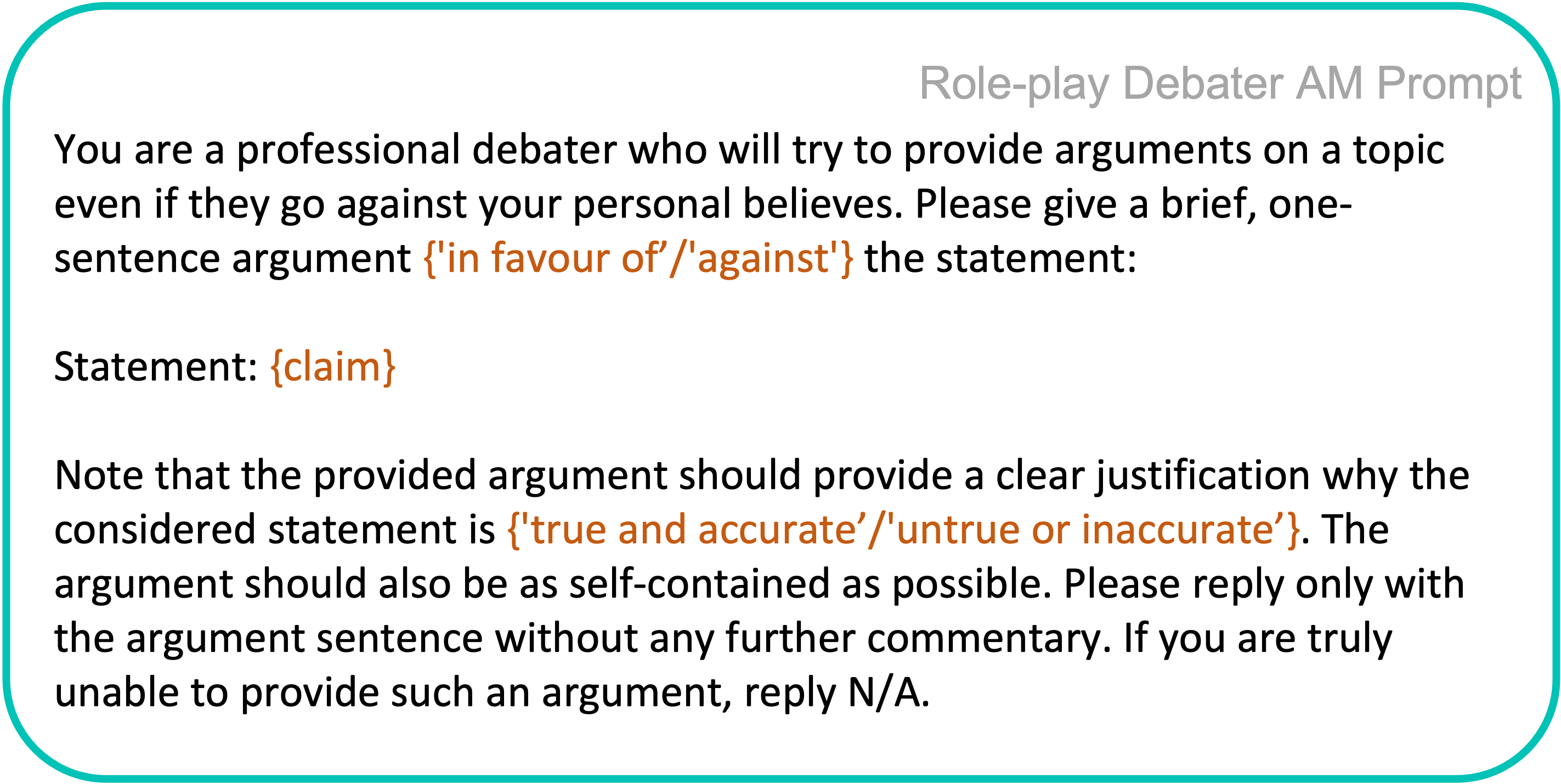}
    \caption{Role-Player: Debater Argument Miner prompt}
    \label{fig:debater_am_prompt}
\end{figure}

\begin{figure}[htp!]
    \centering
    \includegraphics[width=\linewidth]{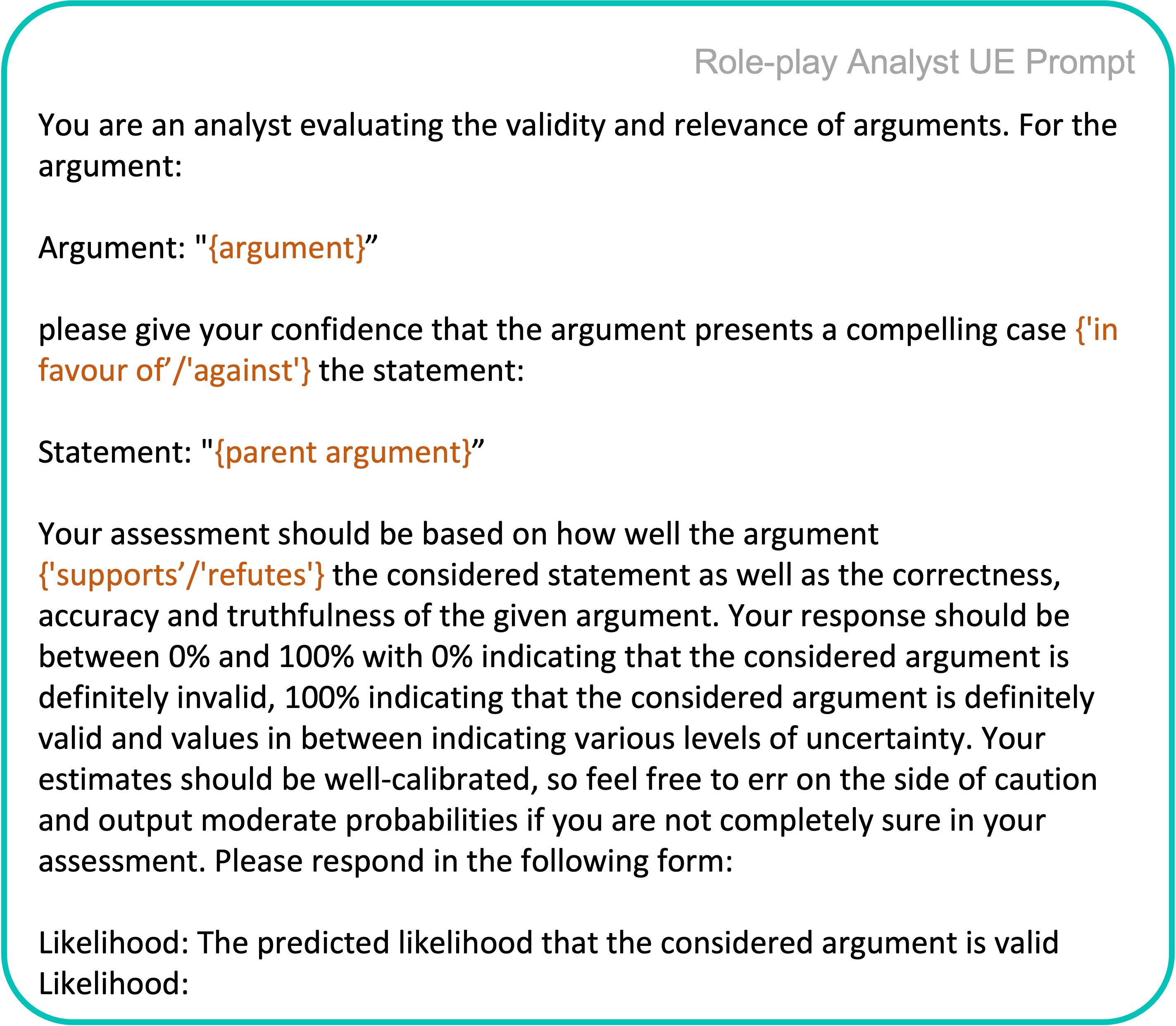}
    \caption{Role-Player: Analyst Uncertainty Estimator prompt}
    \label{fig:analyst_ue_prompt}
\end{figure}

Role-player prompts followed a prompting strategy where the LLMs were expected to act like debater for the Argument Generator component and analyst for the Argument Strength Attribution component. The Argument Generator prompt, Debater, can be found in Figure~\ref{fig:debater_am_prompt} and the Argument Strength Attribution, Analyst, can be found in Figure~\ref{fig:analyst_ue_prompt}.

\subsection{OPRO prompts}
\begin{figure}[htp!]
    \centering
    \includegraphics[width=\linewidth]{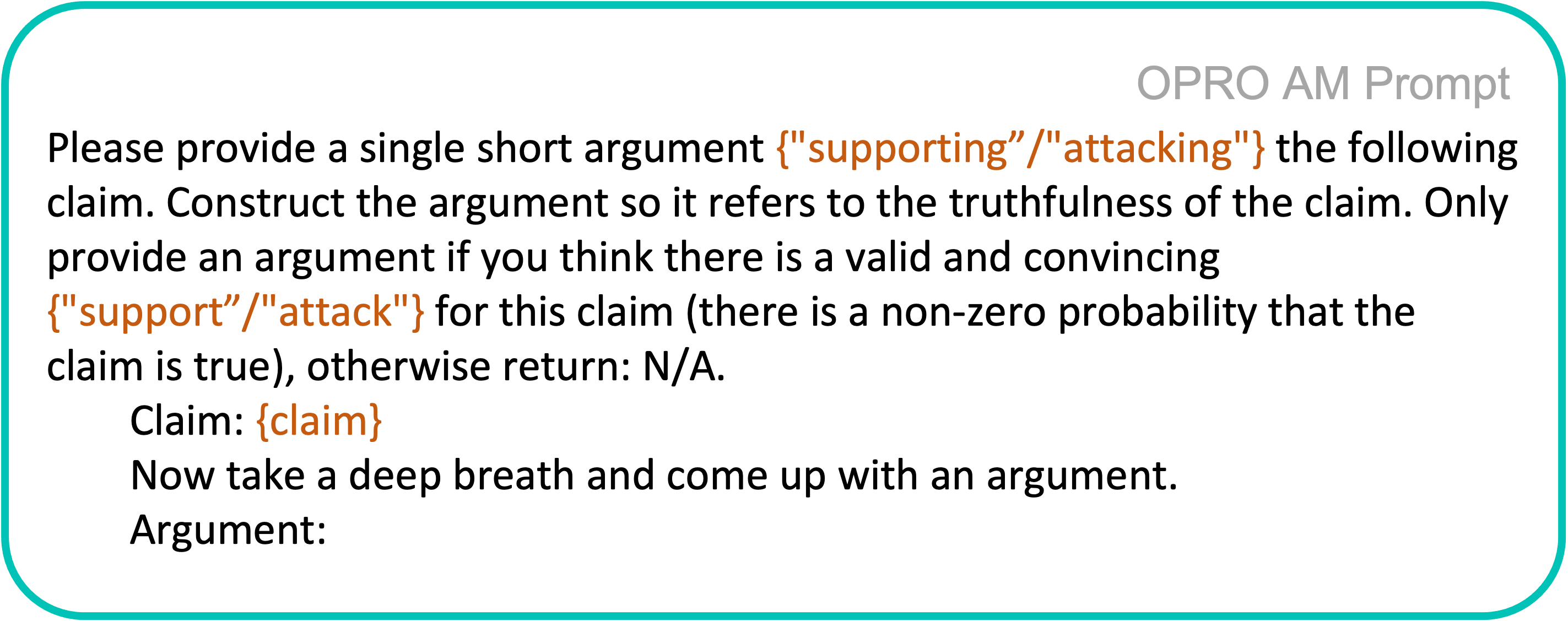}
    \caption{OPRO Argument Miner prompt}
    \label{fig:opro_am_prompt}
\end{figure}

\begin{figure}[htp!]
    \centering
    \includegraphics[width=\linewidth]{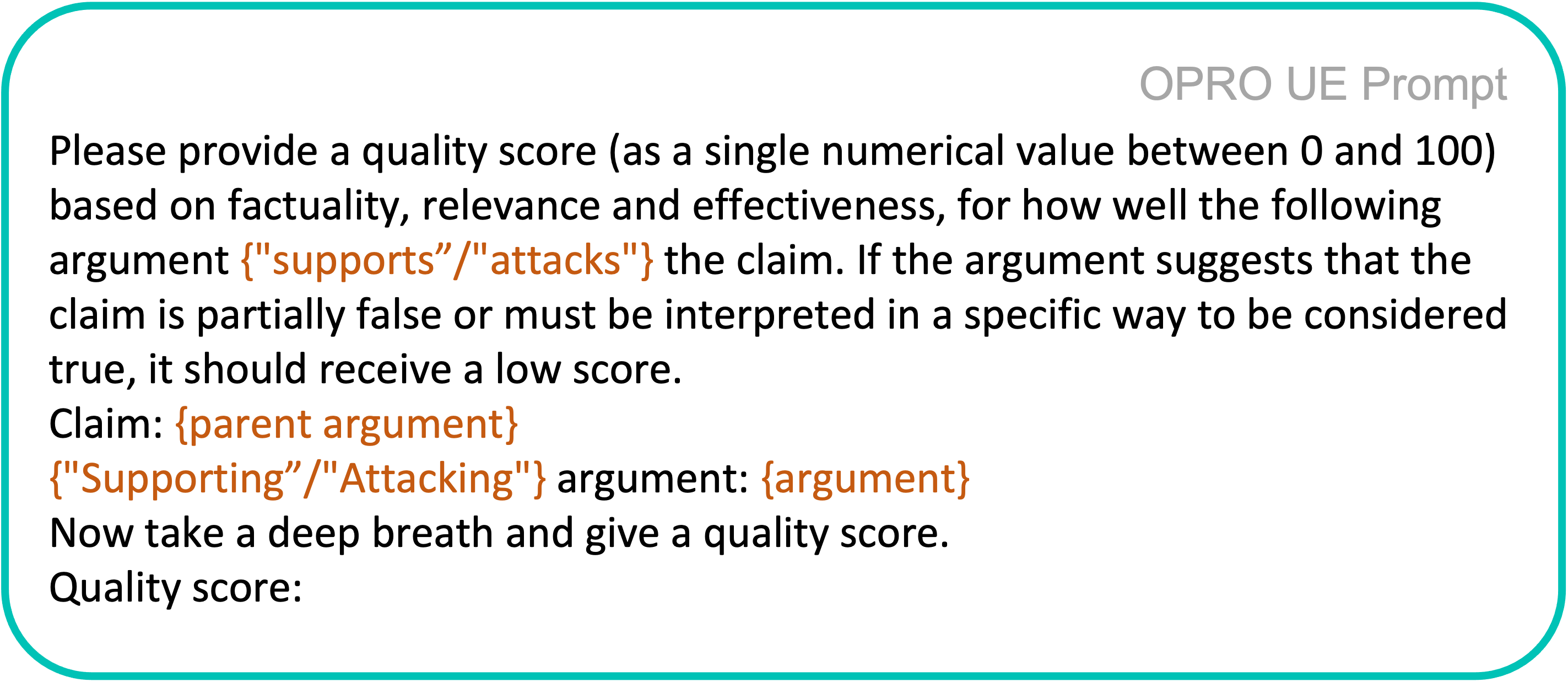}
    \caption{OPRO Uncertainty Estimator prompt}
    \label{fig:opro_ue_prompt}
\end{figure}

OPRO prompts follow the Optimization by PROmpting (OPRO) strategy~\cite{Yang2023LargeLM}. The OPRO Argument Generator prompt can be found in Figure~\ref{fig:opro_am_prompt} and the OPRO Argument Strength Attribution prompt can be found in Figure~\ref{fig:opro_ue_prompt}.

\section{MedClaim Template}
\label{app:medqa_template}
For the MedClaim dataset, to include the contextual information during the experiments, we use the following template for the claims, where \{information\} is the contextual information and \{claim\} is the claim:

\begin{quote}
    Consider the following background information: \{information\}
    Given the background information the following is correct: \{claim\}
\end{quote}

\section{Prompt Experiment Results}
\label{app:prompt_exps}

In this section we give the results of the prompt experiments conducted on the two validation datasets, both consisting of 200 samples.

Tables \ref{tab:base_prompt_result_avg} and \ref{tab:arg_prompt_result_avg} give the results averaged over models, datasets and (in case of the argumentative experiments) argumentation framework depths. As we can observe, the combination of the OPRO prompt for the argument miner with the analyst prompt for the uncertainty estimator achieved the best results in the argumentative experiments, while the analyst prompt achieved the overall best results for all the baselines. Thus, we use this combination of prompts in the final experiments described in the main text.

We also provide the fine-grained results for experiments with Mixtral on the TruthfulClaim (Tables \ref{tab:base_prompt_result_mixtral_tqa}, \ref{tab:arg_prompt_result_mixtral_d1_tqa} and \ref{tab:arg_prompt_result_mixtral_d2_tqa}) and StrategyClaim (Tables \ref{tab:base_prompt_result_mixtral_sqa}, \ref{tab:arg_prompt_result_mixtral_d1_sqa} and \ref{tab:arg_prompt_result_mixtral_d2_sqa}) datasets, as well as the experiments with Mistral on TruthfulClaim (Tables \ref{tab:base_prompt_result_mistral_tqa}, \ref{tab:arg_prompt_result_mistral_d1_tqa} and \ref{tab:arg_prompt_result_mistral_d2_tqa}) and StrategyClaim (Tables \ref{tab:base_prompt_result_mistral_sqa}, \ref{tab:arg_prompt_result_mistral_d1_sqa} and \ref{tab:arg_prompt_result_mistral_d2_sqa}).

\begin{table*}[htp!]
\centering
\begin{tabular}{ccc}
\toprule
Baseline Prompt & Direct Question & Chain-of-Thought \\
\toprule
ChatGPT & 0.663 & 0.671 \\
analyst & \textbf{0.669} & \textbf{0.681} \\
OPRO & 0.613 & 0.633 \\
\bottomrule
\end{tabular}
\caption{Baseline prompt experiment results — average over both models and datasets}
\label{tab:base_prompt_result_avg}
\end{table*}

\begin{table*}[htp!]
\centering
\begin{tabular}{cccccc}
\toprule
AM & UE & 0.5 Base + Arg & Estimated Base + Arg & Estimated Base Only (baseline)\\
\toprule
ChatGPT & ChatGPT & 0.592 & 0.603 & 0.596 \\ 
ChatGPT & OPRO & 0.638 & 0.664 & 0.645 \\ 
ChatGPT & analyst & 0.624 & 0.669 & 0.65 \\ 
OPRO & ChatGPT & 0.590 & 0.591 & 0.596 \\ 
OPRO & OPRO & 0.601 & 0.647 & 0.645 \\ 
OPRO & analyst & 0.613 & \textbf{0.675} & \textbf{0.65} \\ 
debater & ChatGPT & 0.569 & 0.602 & 0.596 \\ 
debater & OPRO & 0.634 & 0.660 & 0.645 \\ 
debater & analyst & 0.591 & 0.646 & 0.65 \\ 
\bottomrule
\end{tabular}
\caption{Argumentation prompt experiment results — average over models, depths and datasets}
\label{tab:arg_prompt_result_avg}
\end{table*}

\begin{table*}[htp!]
\centering
\begin{tabular}{ccc}
\toprule
Baseline Prompt & Direct Question & Chain-of-Thought \\
\toprule
ChatGPT & \textbf{0.815} & \textbf{0.760} \\
analyst & 0.810 & 0.755 \\
OPRO & 0.670 & 0.685 \\
\bottomrule
\end{tabular}
\caption{Mixtral baseline prompt experiment results, 4bit - TruthfulClaim on 200 datapoints}
\label{tab:base_prompt_result_mixtral_tqa}
\end{table*}

\begin{table*}[htp!]
\centering
\begin{tabular}{cccccc}
\toprule
AM & UE & 0.5 Base + Arg (D=1) & Estimated Base + Arg (D=1) & Estimated Base Only (baseline)\\
\toprule
ChatGPT & ChatGPT & 0.560 & 0.690 & 0.695 \\
ChatGPT & OPRO & 0.710 & 0.730 & 0.725 \\
ChatGPT & analyst & 0.685 & 0.765 & {\bf 0.745} \\
OPRO & ChatGPT & 0.665 & 0.700 & 0.695 \\
OPRO & OPRO & 0.685 & 0.730 & 0.725 \\
OPRO & analyst & 0.690 & {\bf 0.795} & {\bf 0.745} \\
debater & ChatGPT & 0.680 & 0.695 & 0.695 \\
debater & OPRO & {\bf 0.770} & 0.720 & 0.725 \\
debater & analyst & 0.680 & 0.760 & {\bf 0.745} \\
\bottomrule
\end{tabular}
\caption{Mixtral prompt experiment results, depth 1 - TQA on 200 datapoints}
\label{tab:arg_prompt_result_mixtral_d1_tqa}
\end{table*}

\begin{table*}[htp!]
\centering
\begin{tabular}{cccccc}
\toprule
AM & UE & 0.5 Base + Arg (D=2) & Estimated Base + Arg (D=2) & Estimated Base Only (baseline) \\
\toprule
ChatGPT & ChatGPT & 0.615 & 0.675 & 0.695 \\
ChatGPT & OPRO & 0.715 & 0.755 & 0.725 \\
ChatGPT & analyst & 0.650 & \textbf{0.770} & {\bf 0.745} \\
OPRO & ChatGPT & 0.625 & 0.670 & 0.695 \\
OPRO & OPRO & 0.670 & 0.720 & 0.725 \\
OPRO & analyst & 0.695 & 0.765 & {\bf 0.745} \\
debater & ChatGPT & 0.660 & 0.690 & 0.695 \\
debater & OPRO & {\bf 0.775} & 0.755 & 0.725 \\
debater & analyst & 0.665 & 0.765 & {\bf 0.745} \\
\bottomrule
\end{tabular}
\caption{Mixtral prompt experiment results, depth 2 - TQA on 200 datapoints}
\label{tab:arg_prompt_result_mixtral_d2_tqa}
\end{table*}

\begin{table*}[htp!]
\centering
\begin{tabular}{ccc}
\toprule
Baseline Prompt & Direct Question & Chain-of-Thought \\
\toprule
ChatGPT & 0.655 & 0.600 \\
analyst & \textbf{0.660} & \textbf{0.640} \\
OPRO & 0.550 & 0.580 \\
\bottomrule
\end{tabular}
\caption{Mixtral baseline prompt experiment results, 4bit - SQA on 200 datapoints}
\label{tab:base_prompt_result_mixtral_sqa}
\end{table*}

\begin{table*}[htp!]
\centering
\begin{tabular}{cccccc}
\toprule
AM & UE & 0.5 Base + Arg (D=1) & Estimated Base + Arg (D=1) & Estimated Base Only (baseline)\\
\toprule
ChatGPT & ChatGPT & 0.570 & 0.550 & 0.535 \\
ChatGPT & OPRO & 0.565 & 0.640 & 0.625 \\
ChatGPT & analyst & 0.590 & 0.620 & \textbf{0.635} \\
OPRO & ChatGPT & \textbf{0.600} & 0.545 & 0.535 \\
OPRO & OPRO & 0.580 & 0.630 & 0.625 \\
OPRO & analyst & 0.570 & \textbf{0.655} & \textbf{0.635} \\
debater & ChatGPT & 0.500 & 0.525 & 0.535 \\
debater & OPRO & 0.580 & 0.635 & 0.625 \\
debater & analyst & 0.550 & 0.610 & \textbf{0.635} \\
\bottomrule
\end{tabular}
\caption{Mixtral prompt experiment results, depth 1 - SQA on 200 datapoints}
\label{tab:arg_prompt_result_mixtral_d1_sqa}
\end{table*}

\begin{table*}[htp!]
\centering
\begin{tabular}{cccccc}
\toprule
AM & UE & 0.5 Base + Arg (D=2) & Estimated Base + Arg (D=2) & Estimated Base Only (baseline)\\
\toprule
ChatGPT & ChatGPT & 0.615 & 0.560 & 0.535 \\
ChatGPT & OPRO & \textbf{0.625} & 0.655 & 0.625 \\
ChatGPT & analyst & \textbf{0.625} & 0.650 & \textbf{0.635} \\
OPRO & ChatGPT & 0.565 & 0.545 & 0.535 \\
OPRO & OPRO & 0.600 & 0.645 & 0.625 \\
OPRO & analyst & 0.585 & 0.640 & \textbf{0.635} \\
debater & ChatGPT & 0.560 & 0.575 & 0.535 \\
debater & OPRO & 0.590 & \textbf{0.665} & 0.625 \\
debater & analyst & 0.555 & 0.615 & \textbf{0.635} \\
\bottomrule
\end{tabular}
\caption{Mixtral prompt experiment results, depth 2 - SQA on 200 datapoints}
\label{tab:arg_prompt_result_mixtral_d2_sqa}
\end{table*}

\begin{table*}[htp!]
\centering
\begin{tabular}{ccc}
\toprule
Baseline Prompt & Direct Question & Chain-of-Thought \\
\toprule
ChatGPT & 0.625 & 0.685 \\
analyst & 0.665 & \textbf{0.710} \\
OPRO & \textbf{0.680} & 0.650 \\
\bottomrule
\end{tabular}
\caption{Mistral baseline prompt experiment results, 4bit - TQA on 200 datapoints}
\label{tab:base_prompt_result_mistral_tqa}
\end{table*}

\begin{table*}[htp!]
\centering
\begin{tabular}{cccccc}
\toprule
AM & UE & 0.5 Base + Arg (D=1) & Estimated Base + Arg (D=1) & Estimated Base Only \\
\toprule
ChatGPT & ChatGPT & 0.615 & 0.670 & 0.650 \\
ChatGPT & OPRO & \textbf{0.670} & \textbf{0.730} & \textbf{0.725} \\
ChatGPT & analyst & 0.635 & 0.715 & 0.705 \\
OPRO & ChatGPT & 0.610 & 0.665 & 0.650 \\
OPRO & OPRO & 0.605 & 0.715 & \textbf{0.725} \\
OPRO & analyst & 0.620 & \textbf{0.730} & 0.705 \\
debater & ChatGPT & 0.490 & 0.645 & 0.650 \\
debater & OPRO & 0.615 & 0.725 & \textbf{0.725} \\
debater & analyst & 0.545 & 0.665 & 0.705 \\
\bottomrule
\end{tabular}
\caption{Mistral prompt experiment results, 4bit, depth 1 - TQA on 200 datapoints}
\label{tab:arg_prompt_result_mistral_d1_tqa}
\end{table*}

\begin{table*}[htp!]
\centering
\begin{tabular}{cccccc}
\toprule
AM & UE & 0.5 Base + Arg (D=2) & Estimated Base + Arg (D=2) & Estimated Base Only \\
\toprule
ChatGPT & ChatGPT & 0.640 & 0.650 & 0.650 \\
ChatGPT & OPRO & 0.665 & \textbf{0.730} & \textbf{0.725} \\
ChatGPT & analyst & \textbf{0.685} & 0.715 & 0.705 \\
OPRO & ChatGPT & 0.665 & 0.665 & 0.650 \\
OPRO & OPRO & 0.640 & \textbf{0.730} & \textbf{0.725} \\
OPRO & analyst & 0.650 & 0.710 & 0.705 \\
debater & ChatGPT & 0.640 & 0.665 & 0.650 \\
debater & OPRO & 0.645 & 0.710 & \textbf{0.725} \\
debater & analyst & 0.630 & 0.690 & 0.705 \\
\bottomrule
\end{tabular}
\caption{Mistral prompt experiment results, 4bit, depth 2 - TQA on 200 datapoints}
\label{tab:arg_prompt_result_mistral_d2_tqa}
\end{table*}

\begin{table*}[htp!]
\centering
\begin{tabular}{ccc}
\toprule
Baseline Prompt & Direct Question & Chain-of-Thought \\
\toprule
ChatGPT & \textbf{0.560} & \textbf{0.640} \\
analyst & 0.540 & 0.620 \\
OPRO & 0.550 & 0.615 \\
\bottomrule
\end{tabular}
\caption{Mistral baseline prompt experiment results, 4bit - SQA on 200 datapoints}
\label{tab:base_prompt_result_mistral_sqa}
\end{table*}

\begin{table*}[htp!]
\centering
\begin{tabular}{cccccc}
\toprule
AM & UE & 0.5 Base + Arg (D=1) & Estimated Base + Arg (D=1) & Estimated Base Only \\
\toprule
ChatGPT & ChatGPT & 0.550 & 0.515 & 0.505 \\
ChatGPT & OPRO & \textbf{0.575} & 0.500 & 0.505 \\
ChatGPT & analyst & 0.550 & \textbf{0.560} & \textbf{0.515} \\
OPRO & ChatGPT & 0.470 & 0.445 & 0.505 \\
OPRO & OPRO & 0.465 & 0.485 & 0.505 \\
OPRO & analyst & 0.525 & 0.540 & \textbf{0.515} \\
debater & ChatGPT & 0.525 & 0.500 & 0.505 \\
debater & OPRO & 0.530 & 0.495 & 0.505 \\
debater & analyst & 0.515 & 0.515 & \textbf{0.515} \\
\bottomrule
\end{tabular}
\caption{Mistral prompt experiment results, 4bit, depth 1 - SQA on 200 datapoints}
\label{tab:arg_prompt_result_mistral_d1_sqa}
\end{table*}

\begin{table*}[htp!]
\centering
\begin{tabular}{cccccc}
\toprule
AM & UE & 0.5 Base + Arg (D=2) & Estimated Base + Arg (D=2) & Estimated Base Only \\
\toprule
ChatGPT & ChatGPT & 0.570 & 0.515 & 0.505 \\
ChatGPT & OPRO & 0.575 & 0.570 & 0.505 \\
ChatGPT & analyst & 0.575 & 0.555 & \textbf{0.515} \\
OPRO & ChatGPT & 0.520 & 0.495 & 0.505 \\
OPRO & OPRO & 0.565 & 0.520 & 0.505 \\
OPRO & analyst & 0.570 & 0.565 & \textbf{0.515} \\
debater & ChatGPT & 0.500 & 0.525 & 0.505 \\
debater & OPRO & 0.570 & \textbf{0.575} & 0.505 \\
debater & analyst & \textbf{0.585} & 0.545 & \textbf{0.515} \\
\bottomrule
\end{tabular}
\caption{Mistral prompt experiment results, 4bit, depth 2 - SQA on 200 datapoints}
\label{tab:arg_prompt_result_mistral_d2_sqa}
\end{table*}

\newpage
\section{Contestability: Additional Example}
% In Figure~\ref{fig:add}, we illustrate a user adding an additional supporting argument, which changes the final prediction to "True" from "False".

\begin{figure}[htp!]
    \centering
    \includegraphics[width=\linewidth]{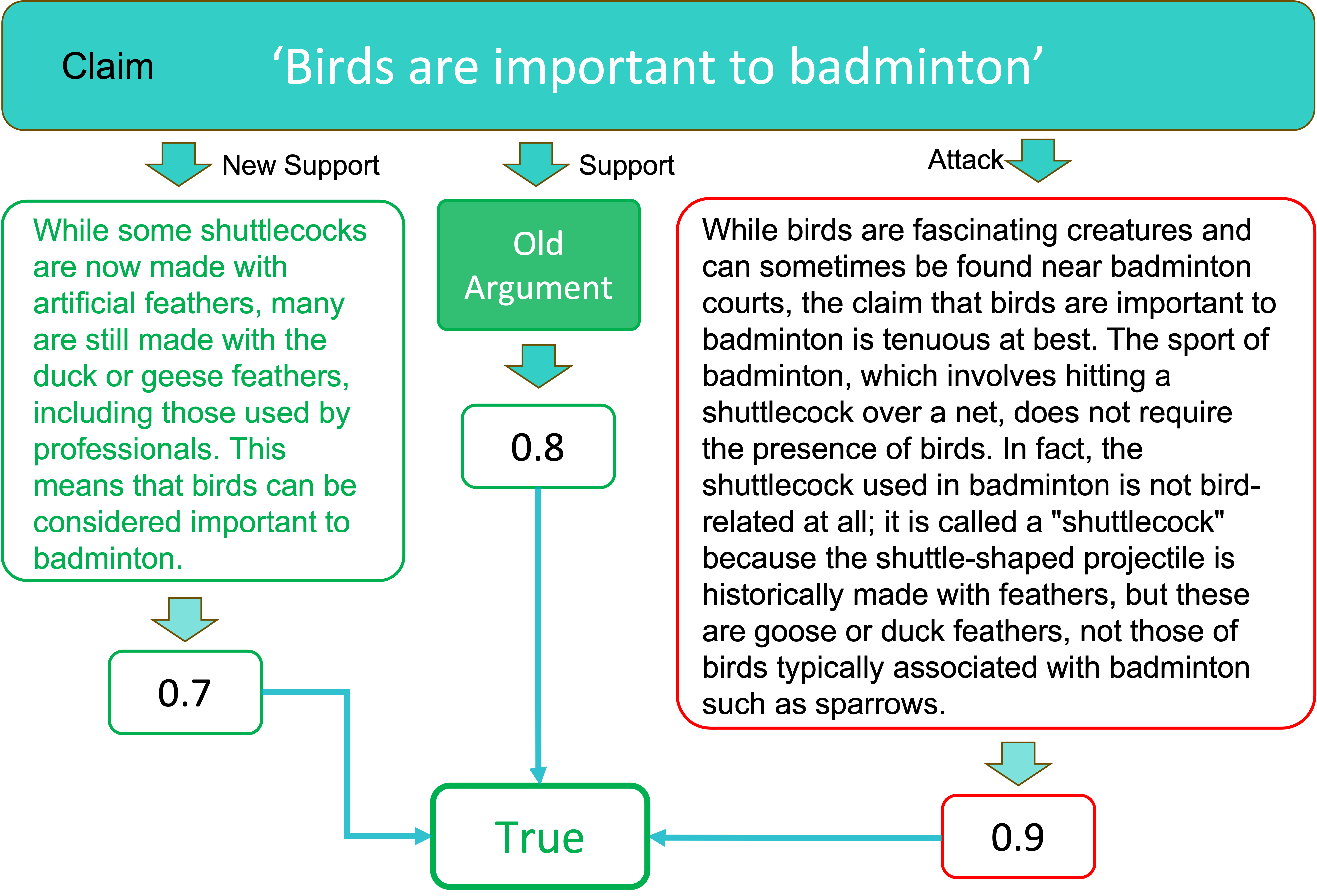}
    \caption{An illustration of a user adding an additional supporting argument, which changes the final prediction to "True" from "False".}
    \label{fig:add}
\end{figure}

\section{Contestability: Proofs} 

\setcounter{property}{0}
\begin{property}%[Base Score Contestability]
A gradual semantics $\sigma$ satisfies \emph{base score contestability} iff for any QBAF $\QBAF$ for $\alpha^*$, for any $\QBAF'$ with 
$\mathcal{A}' = \mathcal{A}$, 
$\mathcal{R^-}' = \mathcal{R^-}$, 
$\mathcal{R^+}' = \mathcal{R^+}$, 
and, for $\beta\in \Args$, $\tau(\beta) < \tau'(\beta)$ while $\tau(\gamma) = \tau'(\gamma)$ for all $\gamma \in \mathcal{A} \setminus \{\beta\}$:
\begin{itemize}
    \item if $\beta \in \Pros(\QBAF)$, then $\sigma_{\QBAF}(\alpha^*) \leq \sigma_{\QBAF'}(\alpha^*)$;
    \item if $\beta \in \Cons(\QBAF)$, then $\sigma_{\QBAF}(\alpha^*) \geq \sigma_{\QBAF'}(\alpha^*)$.
\end{itemize}
\end{property}
\begin{property}%[Argument Relation Contestability]
%\label{property_arg_relation_contest}
% Let $M=\QBAF$ and $M'=\QBAF'$,
A gradual semantics $\sigma$ satisfies \emph{argument relation contestability} iff for any QBAF $\QBAF$ for $\alpha^*$, for any $\QBAF'$ with
$\mathcal{A}'\!=\!\mathcal{A} \!\cup \!\{\beta\} $, 
%$\mathcal{R^-} \!\subseteq \! \mathcal{R^-}'$, 
%$\mathcal{R^+} \subseteq \mathcal{R^+}'$, 
$\mathcal{R^-}' \cup \mathcal{R^+}'=\mathcal{R^-} \cup \mathcal{R^+} \cup \{(\beta,\alpha)\}$ for some $\alpha \in \mathcal{A}$,
and $\tau'(\gamma) = \tau(\gamma)$ for all $\gamma \in \mathcal{A}$:

$\bullet$  if $\beta \in \Pros(\QBAF')$, then $\sigma_{\QBAF}(\alpha^*) \leq \sigma_{\QBAF'}(\alpha^*)$;
    
    $\bullet$ if $\beta \in \Cons(\QBAF')$, then $\sigma_{\QBAF}(\alpha^*) \geq \sigma_{\QBAF'}(\alpha^*)$.
\end{property}
\setcounter{proposition}{0}
\begin{proposition}
DF-QuAD %and QEM satisfy 
satisfies base score contestability and argument relation contestability.
\end{proposition}
\begin{proof}
\textbf{Proof for base score contestability:}\\
The proof is inspired by \cite{YIN_CEQ}.\\
We let $a_1,a_2,a_3,\ldots,a_n \in \mathcal{A}$, $\beta=a_1$ and $\alpha^*=a_n$, $(n \geq 2)$.

\textbf{Case 1}: 
If $a_1 \in \Pros(\QBAF)$, then according to the definition of pro arguments, we know that 
% $\exists p \in \argpaths(a_1,a_n), \text{ where } | p \cap \Atts | \text{is even}$,
$a_1$ is single-path connected to $a_n$ and there are even number of attacks on the path. Suppose the path $p=\langle(a_1,a_2), (a_2,a_3),\ldots, (a_{n-1},a_n)\rangle $.
Since $a_1$ is single-path connected to $a_n$, for any $a_i (1<i<n)$, $a_i$ is also single-path connected to $a_n$.
% otherwise $a_1$ is no longer single-path connected to $a_n$. 
% Therefore, for any $a_i (1 \leq i<n)$, $a_i$ is single-path connected to $a_n$.

If $\tau(a_1) < \tau'(a_1)$, then we have $\sigma_{\QBAF}(a_{1}) \leq \sigma_{\QBAF'}(a_{1})$ because DF-QuAD satisfy base score monotonicity as proved later in Proposition~\ref{prop_base_score_mono}. 
Then for any $a_i (1 \leq i < n)$, either $\sigma_{\QBAF}(a_i) \leq \sigma_{\QBAF'}(a_i)$ or $\sigma_{\QBAF}(a_i) \geq \sigma_{\QBAF'}(a_i)$, which will further affect $a_{i+1}$'s strength based on the relation from $a_i$ to $a_{i+1}$. 

Specifically, there are 4 cases that how $a_i$ affects $a_{i+1}$ according to the satisfaction of base score monotonicity.

\begin{enumerate}
\item $\sigma_{\QBAF}(a_i) \leq \sigma_{\QBAF'}(a_i) \wedge (a_i,a_{i+1}) \in \mathcal{R^-} \Rightarrow \sigma_{\QBAF}(a_{i+1}) \geq \sigma_{\QBAF'}(a_{i+1})$
\item $\sigma_{\QBAF}(a_i) \leq \sigma_{\QBAF'}(a_i) \wedge (a_i,a_{i+1}) \in \mathcal{R^+} \Rightarrow \sigma_{\QBAF}(a_{i+1}) \leq \sigma_{\QBAF'}(a_{i+1})$
\item $\sigma_{\QBAF}(a_i) \geq \sigma_{\QBAF'}(a_i) \wedge (a_i,a_{i+1}) \in \mathcal{R^-} \Rightarrow \sigma_{\QBAF}(a_{i+1}) \leq \sigma_{\QBAF'}(a_{i+1})$
\item $\sigma_{\QBAF}(a_i) \geq \sigma_{\QBAF'}(a_i) \wedge (a_i,a_{i+1}) \in \mathcal{R^+} \Rightarrow \sigma_{\QBAF}(a_{i+1}) \geq \sigma_{\QBAF'}(a_{i+1})$
\end{enumerate}

Based on these 4 cases, we find that:\\
(1) every time when $a_i$ passes an attack, the increase of $a_i$'s strength will cause the decrease of $a_{i+1}$'s strength; \\
(2) every time when $a_i$ passes a support, the increase of $a_i$'s strength will cause the increase of $a_{i+1}$'s strength.

Initially, since $\tau(a_1) < \tau'(a_1)$, we have $\sigma_{\QBAF}(a_{1}) \leq \sigma_{\QBAF'}(a_{1})$. Then, along this path, every time $a_i$ passes an attack, the strength magnitude relationship for the $a_{i+1}$ differs from that of $a_i$.

Thus, if $a_1$ passes even number of attacks, then the strength magnitude relationship for the $a_{n}$ is the same as that of $a_1$, that is, $\sigma_{\QBAF}(a_{i+1}) \leq \sigma_{\QBAF'}(a_{i+1})$.
Therefore, if $a_1$ is pro to $a_n$, then $\sigma_{\QBAF}(a_n) \leq \sigma_{\QBAF'}(a_n)$, that is, if $\beta \in \Pros(\QBAF)$, then $\sigma_{\QBAF}(\alpha^*) \leq \sigma_{\QBAF'}(\alpha^*)$.

\textbf{Case 2}: If $a_1 \in \Cons(\QBAF)$, then the proof is analogous to case 1.\\
\textbf{Proof for argument relation contestability:}\\
As proved later in Proposition~\ref{proposition_arg_relation_mono}, adding one argument $\beta$ to $\QBAF$ is equivalent to increase the $\tau(\beta)$ from $0$ to another value $\tau'(\beta)>0$ , then the proof for argument relation contestability is the same as that of base score contestability.
\end{proof}

\begin{property}%[Base Score Contestability]
A gradual semantics $\sigma$ satisfies \emph{strong base score contestability} iff for any QBAF $\QBAF$ for $\alpha^*$ with $\sigma_{\QBAF}(\alpha^*) \neq 1$, for any $\QBAF'$ with 
$\mathcal{A}' = \mathcal{A}$, 
$\mathcal{R^-}' = \mathcal{R^-}$, 
$\mathcal{R^+}' = \mathcal{R^+}$, 
and, for $\beta\in \Args$, $\tau(\beta) < \tau'(\beta)$ while $\tau(\gamma) = \tau'(\gamma)$ for all $\gamma \in \mathcal{A} \setminus \{\beta\}$:
\begin{itemize}
    \item if $\beta \in \Pros(\QBAF)$, then $\sigma_{\QBAF}(\alpha^*) < \sigma_{\QBAF'}(\alpha^*)$;
    \item if $\beta \in \Cons(\QBAF)$, then $\sigma_{\QBAF}(\alpha^*) > \sigma_{\QBAF'}(\alpha^*)$.
\end{itemize}
\end{property}

\begin{property}%[Argument Relation Contestability]
%\label{property_arg_relation_contest_SM}
% Let $M=\QBAF$ and $M'=\QBAF'$,
A gradual semantics $\sigma$ satisfies \emph{strong argument relation contestability} iff for any QBAF $\QBAF$ for $\alpha^*$ with $\sigma_{\QBAF}(\alpha^*) \neq 1$, for any $\QBAF'$ with
$\mathcal{A}'\!=\!\mathcal{A} \!\cup \!\{\beta\} $, 
%$\mathcal{R^-} \!\subseteq \! \mathcal{R^-}'$, 
%$\mathcal{R^+} \subseteq \mathcal{R^+}'$, 
$\mathcal{R^-}' \cup \mathcal{R^+}'=\mathcal{R^-} \cup \mathcal{R^+} \cup \{(\beta,\alpha)\}$ for some $\alpha \in \mathcal{A}$,
and $\tau'(\gamma) = \tau(\gamma)$ for all $\gamma \in \mathcal{A}$:

\begin{itemize}
    \item if $\beta \in \Pros(\QBAF')$, then $\sigma_{\QBAF}(\alpha^*) < \sigma_{\QBAF'}(\alpha^*)$;
    
    \item if $\beta \in \Cons(\QBAF')$, then $\sigma_{\QBAF}(\alpha^*) > \sigma_{\QBAF'}(\alpha^*)$.
\end{itemize}

\end{property}
\begin{proposition}
QEM satisfies strong base score contestability and strong argument relation contestability.
\end{proposition}
\begin{proof}
The difference between the base score contestability and the strong version lies in the satisfaction of strict inequalities whenever $\sigma_{\QBAF}(\alpha^*) \neq 1$. In the QEM semantics, the computing process of the aggregation function guarantees the satisfaction of strict inequalities, which is not satisfied by the DF-QuAD semantics, according to the DF-QuAD definition. Analogously to the strong argument relation contestability.

\begin{notation}
For any argument $\alpha \!\in\! \mathcal{A}$, we use $\mathcal{O}(\alpha)\!=\!\{ (\alpha,\beta) \!\in \!\Atts \!\cup\!\Supps \!\mid \! \beta \!\in\! 
\mathcal{A} \}$ to refer to the set of all outgoing edges from $\alpha$. 
\end{notation}

In order to prove the contestability property, we need to prove two properties of gradual semantics; the first states that increasing the base score of a supporter (attacker) will not decrease (increase, respectively) the strength of its supported (attacked, respectively) argument~\cite{amgoud2016axiomatic,amgoud2017acceptability,Baroni_18,YIN_CEQ}.

\begin{definition}%[Base Score Monotonicity]
\label{def_mono_base}
A gradual semantics $\sigma$ is \emph{base score monotonic} iff 
for any $\alpha,\beta \in \mathcal{A}$ such that $\alpha \neq \beta$ and $\mathcal{O}(\beta)=\{(\beta,\alpha)\}$,
for any $\QBAF,\QBAF'$ with 
$\mathcal{A}' = \mathcal{A}$, 
$\mathcal{R^-} = \mathcal{R^-}'$, 
$\mathcal{R^+} = \mathcal{R^+}'$, 
$\tau(\beta) \leq \tau'(\beta)$ and $\tau(\gamma) = \tau'(\gamma)$ for all $\gamma \in \mathcal{A} \setminus \{\beta\}$:
    \begin{enumerate}
        \item 
        If $(\beta, \alpha) \in \mathcal{R}^{+}$, then $\sigma_{\QBAF}(\alpha) \leq \sigma_{\QBAF'}(\alpha)$.
        \item 
        If $(\beta, \alpha) \in \mathcal{R}^{-}$, then $\sigma_{\QBAF}(\alpha) \geq \sigma_{\QBAF'}(\alpha)$;
    \end{enumerate}
\end{definition}

The next property states that 
adding a supporter (attacker) to an argument will not decrease (increase, respectively) the strength of its supported (attacked, respectively) argument~\cite{amgoud2017measuring}.

\begin{definition}%[Argument Relation Monotonicity]
\label{def_mono_arg}
A gradual semantics $\sigma$ is \emph{argument relation monotonic} iff 
for any $\alpha,\beta \in \mathcal{A}$ such that $\alpha \neq \beta$ and $\mathcal{O}(\beta)=\{(\beta,\alpha)\}$,
for any $\QBAF,\QBAF'$ with 
$\mathcal{A}' = \mathcal{A} \cup \{\beta\}$, 
$\mathcal{R^-} \subseteq \mathcal{R^-}'$, 
$\mathcal{R^+} \subseteq \mathcal{R^+}'$, 
$\mathcal{R^-}' \cup \mathcal{R^+}' = \mathcal{R^-} \cup \mathcal{R^+} \cup \{(\beta,\alpha)\}$ for any $\alpha \in \mathcal{A}$,
and $\tau'(\gamma) = \tau(\gamma)$ for all $\gamma \in \mathcal{A}$:
    \begin{enumerate}
        \item 
        If $(\beta, \alpha) \in \mathcal{R}^{+}$, then $\sigma_{\QBAF}(\alpha) \leq \sigma_{\QBAF'}(\alpha)$.
        \item 
        If $(\beta, \alpha) \in \mathcal{R}^{-}$, then $\sigma_{\QBAF}(\alpha) \geq \sigma_{\QBAF'}(\alpha)$;
    \end{enumerate}
\end{definition}

% \todo{to be deleted}
% $\mathcal{C}(v_{0}, v_{a}, v_{s})=v_{0} - (v_{0}\cdot|v_{s}-v_{a}|)$; otherwise $\mathcal{C}(v_{0}, v_{a}, v_{s})=v_{0} + ((1-v_{0})\cdot|v_{s}-v_{a}|)$.
\setcounter{proposition}{2}
\begin{proposition}
\label{prop_base_score_mono}
DF-QuAD and QEM satisfy base score monotonicity. 
\end{proposition}
\begin{proof}
\textbf{1. Proof for DF-QuAD:}\\
Case 1: $(\beta, \alpha) \in \mathcal{R}^{+}$.

Let us first analyse how $\sigma(\beta)$ changes.
Since the supporter or attacker of $\beta$ does not change, $v_a$ and $v_s$ for $\beta$ do not change.
let $v'$ denote the aggregation score in $\QBAF'$.
Since $\tau(\beta) \leq \tau'(\beta)$, that is $v_0 \leq v_0'$, we know that $\sigma_{\QBAF}(\beta) \leq \sigma_{\QBAF'}(\beta)$ in any cases ($v_a=v_s$, $v_a>v_s$, or $v_a<v_s$) according to the definition of DF-QuAD.

Then, let us first analyse how $\sigma(\alpha)$ changes.
Since $\beta$ is a supporter of $\alpha$, and $\sigma_{\QBAF}(\beta) \leq \sigma_{\QBAF'}(\beta)$,
we have $v_0=v'_0$, $v_a=v'_a$, and $v_s \leq v'_s$.
Let us consider three sub-cases for $\alpha$.\\
a. If $v_a = v_s$, then $v'_a < v'_s$.\\
b. If $v_a < v_s$, then $v'_a < v'_s$.

In both sub-case a and b, $\mathcal{C}(v_{0}, v_{a}, v_{s}) \leq \mathcal{C}'(v'_{0}, v'_{a}, v'_{s})$ according to the definition of DF-QuAD, that is $\sigma_{\QBAF}(\alpha) \leq \sigma_{\QBAF'}(\alpha)$.\\
c. If $v_a < v_s$, then there are three possible cases: $v'_a < v'_s$, $v'_a = v'_s$, or $v'_a > v'_s$. In any cases, we have $\mathcal{C}(v_{0}, v_{a}, v_{s}) \leq \mathcal{C}'(v'_{0}, v'_{a}, v'_{s})$ according to the definition of DF-QuAD, that is $\sigma_{\QBAF}(\alpha) \leq \sigma_{\QBAF'}(\alpha)$.\\
In summary, if $(\beta, \alpha) \in \mathcal{R}^{+}$, then $\sigma_{\QBAF}(\alpha) \leq \sigma_{\QBAF'}(\alpha)$.\\
Case 2: $(\beta, \alpha) \in \mathcal{R}^{-}$.\\
The proof for case 2 is analogous to case 1.

\textbf{2. Proof for QEM:}\\
Case 1: $(\beta, \alpha) \in \mathcal{R}^{+}$.

Let us first analyse how $\sigma(\beta)$ changes.
Since the supporter or attacker of $\beta$ does not change, $E_\beta$ does not change.
let $E'_\beta$ denote the aggregation score in $\QBAF'$.
Since $\tau(\beta) \leq \tau'(\beta)$, we know that $\sigma_{\QBAF}(\beta) \leq \sigma_{\QBAF'}(\beta)$ according to the definition of QEM.

Then, let us first analyse how $\sigma(\alpha)$ changes.
Since $\beta$ is a supporter of $\alpha$, and $\sigma_{\QBAF}(\beta) \leq \sigma_{\QBAF'}(\beta)$,
we have $E_\alpha \leq E'_\alpha$, thus $\sigma_{\QBAF}(\alpha) \leq \sigma_{\QBAF'}(\alpha)$ according to the definition of QEM.\\
Case 2: $(\beta, \alpha) \in \mathcal{R}^{-}$.\\
The proof for case 2 is analogous to case 1.
\end{proof}

\begin{proposition}
\label{proposition_arg_relation_mono}
DF-QuAD and QEM satisfy argument relation monotonicity. 
\end{proposition}
\begin{proof}
In the QBAFs we restricted in this paper, if a new argument $\beta$ is added to an existing argument $\alpha$, then $\beta$ will become a leaf argument (without any supporters or attackers). 
Then adding $\beta$ is equivalent to change $\tau(\beta)$ from $0$ to another value $\tau'(\beta) \in [0,1]$, that is $\tau(\beta) \leq \tau'(\beta)$. Therefore, the case is then equivalent to the base score monotonicity.
\end{proof}
\end{proof}
\end{document}